\documentclass[11pt]{article}

\usepackage[margin=1in]{geometry}
\usepackage[utf8]{inputenc}
\usepackage[T1]{fontenc}
\usepackage{amsmath,amssymb,amsfonts}
\usepackage{graphicx}
\usepackage{float}
\usepackage{booktabs}
\usepackage{multirow}
\usepackage{xcolor}
\usepackage{xspace}
\usepackage[colorlinks=true,linkcolor=blue,citecolor=blue,urlcolor=blue]{hyperref}
\usepackage[capitalize,noabbrev]{cleveref}
\usepackage{caption}
\usepackage{enumitem}
\usepackage{titlesec}

\newcommand{\method}{WeaveAgent\xspace}
\newcommand{\weaveearth}{WeaveEarth\xspace}
\newcommand{\remoteagent}{RemoteAgent\xspace}
\newcommand{\tcall}{\texttt{T\_call}\xspace}
\newcommand{\mses}{MSES\xspace}
\newcommand{\tpeb}{TPEB\xspace}
\newcommand{\sem}{SEM\xspace}
\newcommand{\gcc}{GCC\xspace}
\newcommand{\vagueuhr}{VagueUHR\xspace}

\title{\textbf{WeaveAgent: A Two-Stage Tool-Routing Agent for\\
Ultra-High-Resolution Remote Sensing Imagery}}

\author{Zhongyu Pang\\[4pt]
  \small\textit{Department of Electronics, National University of Defense Technology,
  Changsha, China}}
\date{September 2026}

\begin{document}
\maketitle

\begin{abstract}
\textbf{Problem.} Ultra-high-resolution (UHR) remote sensing with vague
user intents, the form field analysts actually use when querying
imagery for disaster assessment, urban monitoring, and maritime traffic
control, poses two bottlenecks: \emph{visual tokens are expensive},
and \emph{tool calling must be format-reliable} (pretrained models emit
zero tool calls zero-shot).
\textbf{Method.} \method{}, a two-stage tool-routing agent, decouples
routing from visual perception. \emph{Stage~A} is
routing-first: emission is trained, not elicited. \emph{Stage~B}
executes conditionally: intrinsic queries enter visual answering via an
optional fixed-budget ${\approx}5$k-token compression interface;
extrinsic queries execute \tcall{} on the \emph{original}
full-resolution imagery, answering from tool observations in a second,
observation-masked round. Training: alignment SFT, then GRPO under the
routing-first reward $R_{\mathrm{WA2}}$.
\textbf{Results.} Alignment SFT lifts extrinsic routing from $0\%$ to
\textbf{80.75\%} (323/400); GRPO suppresses 9 intrinsic mis-emissions
while tool selection is unchanged. The contribution is diagnostic rather than
leaderboard-driven, isolating which training stage buys which
capability. Oracle attribution separates
the repair's two ingredients: loading the observation into context
lifts extrinsic answer accuracy from $0.025$ to $0.425$ under
marker-free cross-mode returns, and the two-turn SFT stage adds a
significant $+9.3$ points to $0.518$ ($0.965$ under
training-distribution returns; all oracle/annotation-grounded) at a
small routing cost. A $\pm$-image ablation shows emission suppression
is visually grounded, and a query-register matrix shows LLM-rewritten
queries cost the trained checkpoints $2$--$11$ points.
\textbf{Scope.} All training and evaluation use the
5{,}000\,/\,3{,}273\,/\,1{,}000-record \vagueuhr{} corpus (600 intrinsic
$+$ 400 tool-requiring; the base seeds synthesis and is not used for
optimization). Single-pass evidence construction
runs at 7.31\,s per image (RTX\,4090; cross-hardware reference).
Code, data, and all evaluation protocols will be released.
\end{abstract}


\section{Introduction}
\label{sec:intro}

\begin{figure}[t]
  \centering
  \includegraphics[width=0.98\textwidth]{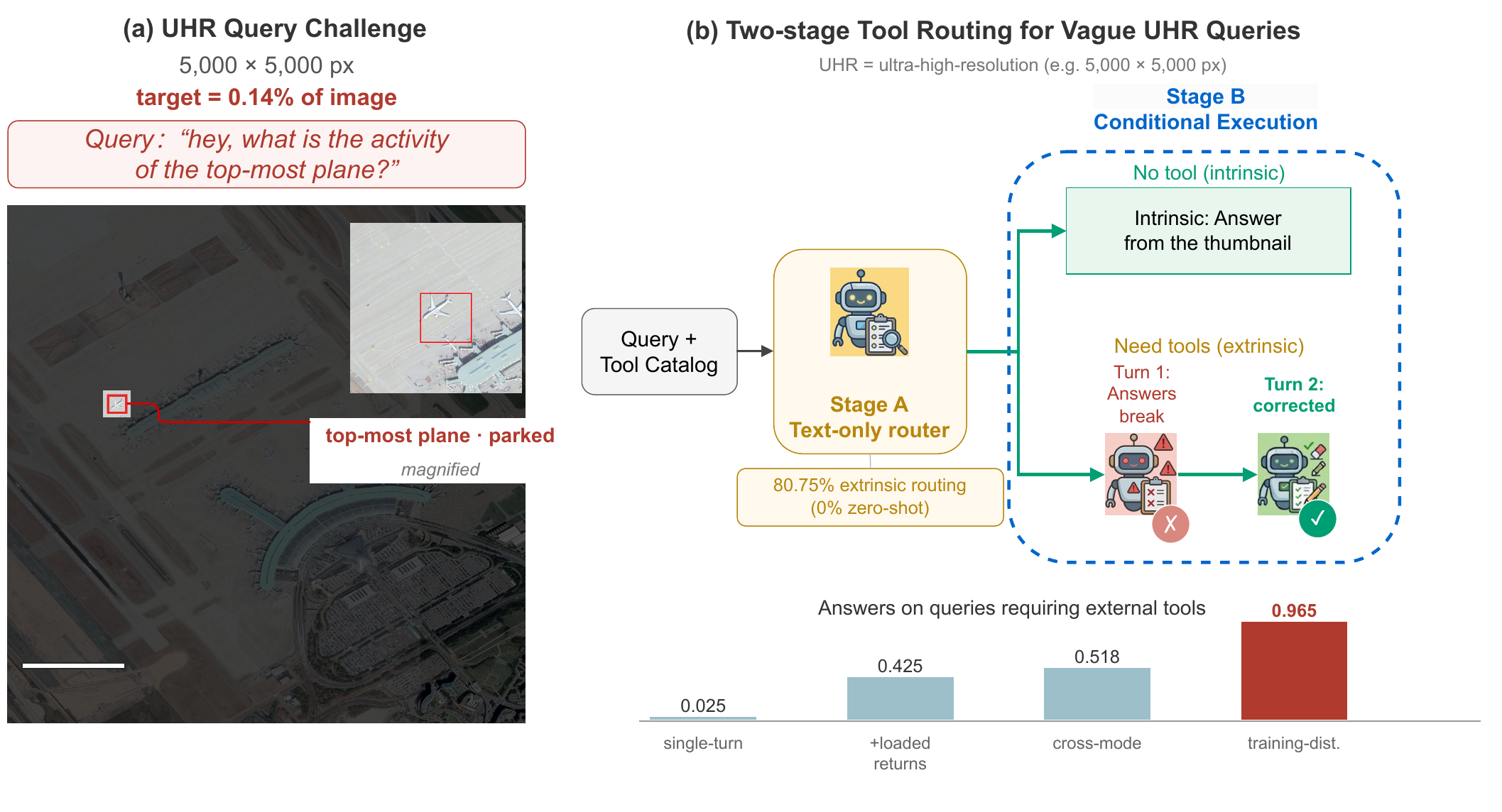}
  \caption{The problem and the answer at a glance. \textbf{Left:} a real
  test record, a vague query about a target that covers $0.14\%$ of a
  $5{,}000\times5{,}000$-px image (magnified inset); no thumbnail a VLM
  accepts resolves it. \textbf{Right:} \method{} routes the query first
  (Stage~A, a routing-first decision stage) and executes
  conditionally (Stage~B): direct visual answering, or a \tcall{} that is
  answered from its observation in a second round. Headline numbers:
  extrinsic routing $0\%\to80.75\%$ after alignment training; answers on
  tool-bearing rows $0.025\to0.425\to0.518\to0.965$ across the loading
  and repair stages
  (\cref{sec:conversion}).}
  \label{fig:teaser}
\end{figure}

Ultra-high-resolution (UHR) remote sensing (RS) images, sub-meter scenes
that routinely exceed 4{,}000--10{,}000 pixels per side, stress
vision-language models (VLMs) in a specific way: resizing the scene to the
model's native input discards exactly the dense small objects that make
UHR imagery informative, while pushing the original resolution through
the visual encoder makes the token cost grow out of control (the images
evaluated in this work measure 6{,}044\,px per side at the median and
27{,}328\,px at the maximum; \cref{fig:teaser} shows both difficulties on
one real record). On top of the visual difficulty, RS queries are
frequently \emph{vague} in the sense of \remoteagent{}
\cite{yao2026remoteagent}: ``count the
berthed vessels'', ``outline the burned area''. Such requests need
precise measurement that no single forward pass can deliver; they call
for expert tools (detectors, segmenters, counters) applied to the right
image region. Berthed-vessel counts and burned-area extents are the
working quantities of disaster-response and urban-monitoring imagery
workflows, the settings in which UHR archives are actually queried.

The two bottlenecks point to two capabilities of a different nature.
\emph{Perception is expensive}: every visual token costs, and should be
spent only after the decision ``what needs to be seen'' has been made.
\emph{Decision is cheap}: the information needed to judge ``does this
query need a tool, and which one'' lives almost entirely in the query
text and the tool catalog. Two lines of work support this split:
Toolformer~\cite{schick2023toolformer}
showed that a language model can learn \emph{when to call, what to call,
and with which arguments} from purely textual self-supervision, and
RouteLLM~\cite{ong2024routellm} showed that a router looking only at the
query text can halve cost without quality loss and generalizes across
models. When these findings are transported to UHR remote sensing, the
natural architecture is to
\emph{decouple routing decisions from visual perception}: a text-first
decision stage decides before any visual token is spent on answering;
visual tokens enter the
context only afterwards, on the path the decision selects. Our own
$\pm$-image ablation then shows that the strong form of this design
basis is only partially attained: the trained router's \emph{emission
suppression} is visually grounded (\cref{sec:pmimage}).

\paragraph{The \method{} two-stage architecture.} Stage~A
(\cref{sec:router}) is a \emph{routing-first decision stage}: its decision
interface is built around the query $q$
and the tool catalog $C$ (a text-first decision interface), delegates intrinsic
queries to answering, and for extrinsic queries selects a tool and emits
$\tcall(e_k, p)$. Stage~B (\cref{sec:intrinsic}--\cref{sec:extrinsic})
executes conditionally: the \emph{intrinsic path} feeds the visual input
(full-scene thumbnail, optionally the evidence board) to the model to
generate the answer; the \emph{extrinsic path} executes the \tcall{}
(tools always operate on the \emph{original} full-resolution imagery),
backfills the tool observation (boxes, masks, counts) into the context,
and generates the final answer in a second round (two-turn trajectories
with observation masking~\cite{openearthagent}). Decoupling yields three
direct benefits: (i) the decision stage is a single cheap text-first step
(\cref{sec:pmimage} quantifies how far a fully text-only router remains
from the trained policy); (ii)
the \tcall{} output is a purely textual decision that alignment training
can make reliably parseable; and (iii) the tool path
suffers no visual-compression loss, because tools see the original image.

\paragraph{Repositioning the evidence board.} We adopt the
\weaveearth{}~\cite{ma2026weaveearth} structured evidence pipeline
(the global context constraint (\gcc{}) scoring, the minimal support evidence set (\mses{})
compression, and the topology-preserving evidence board (\tpeb{}) /
structured evidence metadata block (\sem{}) assembly) as an
\emph{optional compression interface} of the Stage-B intrinsic path, and
recalibrate its position. It is \emph{not} an accuracy booster:
on three static VQA benchmarks its paired deltas against the 2048-px
thumbnail baseline are $-1.0$ (LRS-VQA), $-3.4$ (MME-RS) and $-2.9$
(XLRS) points, and its components show no positive contribution in
ablations (\cref{sec:ablation}). It \emph{is} a fixed-budget compressed
visual interface: any UHR image is compressed into two structured
views of ${\approx}5$k tokens with constant token cost regardless of
source size, trading a $1.31\times$ detail ceiling (\cref{sec:ceiling})
for determinism and local-structure preservation on fine-grained task
types. All of its properties are reported below under this ``compression
interface, not booster'' positioning.

\paragraph{Training recipe: alignment SFT $\rightarrow$ $R_{\mathrm{WA2}}$
GRPO, a positive gain per component.} Tool routing does not appear for
free: the zero-shot model emits zero \tcall{} calls, and GRPO from the
base model cannot explore the behavior spontaneously (85--90\% of rollout
groups have zero reward variance). Our recipe has two steps.
\emph{Alignment SFT} seeds the emission behavior with prompts and image
formats strictly identical to evaluation, lifting extrinsic routing from
$0\%$ to 80.75\% (323/400, tool selection 0.905). \emph{GRPO with the
routing-first reward $R_{\mathrm{WA2}}$} then stabilizes on top:
suppressing 9 intrinsic mis-emissions (classification
$0.435\to0.460$, reasoning $0.435\to0.450$) while extrinsic tool
selection is unchanged (323/400), with well-behaved training
dynamics (zero-variance groups an order of magnitude below the
unaligned arms, sequence lengths stable at
165--180 tokens). Under this recipe a third gap surfaced and was
addressed: with routing at 80.75\%, extrinsic \emph{answer} accuracy
stayed near zero. This is an observation$\to$answer \emph{conversion
gap}. An
oracle attribution study (\cref{sec:conversion}) localizes the break to
the conversion of marker-free tool text (argument values are
${\approx}96\%$ correct while final answers score $0.045$), and
extrinsic answer accuracy rises from 0.025 to 0.425 by loading the
observation alone, and to 0.518 with the two-turn SFT stage (McNemar
$p=3.8\times10^{-5}$), under marker-free cross-mode returns (0.965 under
training-distribution returns), at a small routing cost. Query
\emph{wording} is not a hidden confound: on 992 paired records the
zero-shot floor is robust to LLM-rewritten natural phrasing ($+3.1$
points, McNemar $p=0.043$; \cref{sec:wording}), and the trained
checkpoints degrade gracefully but measurably on the rewritten register
(\cref{sec:register}).

\paragraph{Contributions.}
\begin{enumerate}[leftmargin=*,itemsep=2pt]
  \item \textbf{C1: A two-stage tool-routing agent architecture.} To
  our knowledge, the first agent architecture for UHR remote sensing that
  explicitly decouples a \emph{routing-first} decision stage from
  \emph{conditional visual/tool execution}, with the stage's emission
  behavior trained before any execution is attempted: Stage~A performs
  intrinsic/extrinsic discrimination and tool selection from the query
  text and the catalog; Stage~B answers intrinsic queries visually and
  executes tools on the original full-resolution imagery along two-turn
  trajectories with observation masking. A $\pm$-image ablation shows
  the stage's emission suppression is visually grounded: removing image
  tokens preserves extrinsic routing but collapses intrinsic routing to
  near zero (\cref{sec:pmimage}). The implementation spans six
  subsystems (\texttt{evidence}, \texttt{skills}, \texttt{agents},
  \texttt{mcp}, \texttt{training}, \texttt{data\_synthesis}), twelve
  built-in tools behind a declarative MCP registry~\cite{anthropic2024mcp}
  with automatic discovery, and a publish/subscribe multi-agent layer.
  \item \textbf{C2: A fixed-budget compressed visual interface
  (repositioned).} The \tpeb{}$+$\sem{} evidence board as a compression
  interface for fine-grained task types: any-size UHR image $\to$ fixed
  token budget (two views, ${\approx}5$k tokens), detail ceiling
  $1.31\times$ reported as measured; three-benchmark paired characterization
  ($-1.0/-3.4/-2.9$ points) and component ablations (\cref{sec:ablation})
  jointly support ``compression interface, not accuracy booster''.
  \item \textbf{C3: The \vagueuhr{} dual-split corpus.} A vague-query
  trajectory corpus synthesized over UHR imagery: 5{,}000 template
  training (synthesis base; not used for optimization) / 3{,}273 routing-training (870 tool-call demonstrations) /
  1{,}000 test records (600 intrinsic $+$ 400 tool-requiring; five of ten
  task shapes unseen in training; tool pool includes the training-unseen
  tool sm3det with 62 records); plus an LLM-visually-rewritten test split
  (992 kept $+$ 8 removed; 100\% unique queries, 143 sentence-initial
  word types, zero template phrasing, character-level zero drift of
  ground truth under independent audit) (\cref{sec:data}).
  \item \textbf{C4: A training recipe with design-rationale
  experiments.} The alignment-SFT $\rightarrow$ $R_{\mathrm{WA2}}$ GRPO
  recipe yields a checkable gain concentrated in alignment SFT
  (extrinsic routing $0\to$80.75\%, 323/400), with GRPO contributing a
  measured, statistically modest effect (9 intrinsic mis-emissions
  suppressed, calls $398\to384$; extrinsic tool selection unchanged at
  323/400; Mean $+0.7$pt, n.s.), and the design-rationale experiments of
  \cref{sec:rationale} back each choice with data (why alignment is
  necessary, why the incentive must be routing-first, the boundary of
  the compression interface, and an oracle attribution $+$ two-turn
  repair of the observation$\to$answer conversion gap), each
  transferable to other tool-routing tasks.
\end{enumerate}

\paragraph{Paper organization.} \cref{sec:related} reviews related work,
including text-only routing and cascades. \cref{sec:method} presents the
two-stage architecture (Stage~A, the Stage-B intrinsic and extrinsic
paths, the training recipe, and the two-turn repair). \cref{sec:data}
describes \vagueuhr{}. \cref{sec:expts} reports experiments (the routing
table, compression-interface characterization and ablations, efficiency,
the query-register gradient, the Stage-A
$\pm$-image ablation, and tool-execution conversion).
\cref{sec:rationale} gives the design-rationale analyses.
\cref{sec:limitations} lists limitations and \cref{sec:conclusion}
concludes. Supplementary tables, figures, and the experiment register
are collected in Appendices~A--C.

\section{Related Work}
\label{sec:related}

\paragraph{UHR remote sensing understanding: passive vs.\ active
perception.} \emph{Passive} methods push more pixels or tokens through the
model: GeoLLaVA-8K~\cite{geollava} extends the visual context to 8K, and
token-pruning pipelines compress gigapixel inputs at inference
time~\cite{luo2025lrsvqa}. \emph{Active} methods let the model zoom and
search: ZoomEye~\cite{zoomeye}, ZoomEarth~\cite{zoomearth} and
ZoomSearch~\cite{zoomsearch} perform multi-round visual exploration driven
by the language model, at multi-round latency cost (\cref{sec:efficiency}).
\weaveearth{}~\cite{ma2026weaveearth} reframes the problem as
\emph{evidence organization}: candidate patches scored by a cross-modal
encoder (SigLIP2~\cite{zhai2025siglip2}), greedily compressed into a
minimal support set, and arranged on a topology-preserving board with
structured metadata; it reports 33.38/47.38/47.14\% on LRS-VQA /
MME-RealWorld-RS / XLRS-Bench with a frozen
Qwen3-VL-8B~\cite{bai2025qwen3vl}. All of these are \emph{tool-free}:
whatever they cannot resolve from pixels and prompts, they cannot answer.
\method{} adopts the evidence pipeline as Stage-B's optional compression
interface (\cref{sec:evidence}) and adds full tool-routing capability on
top.

\paragraph{RS vision-language models and instruction tuning.} A parallel
line trains RS-specialized MLLMs by instruction tuning:
GeoChat~\cite{geochat} grounds conversations in RS imagery,
EarthGPT~\cite{earthgpt} and its spatial successor
EarthGPT-X~\cite{earthgptx} unify multisensor, multitask interpretation
under one MLLM, RSGPT~\cite{rsgpt} and
LHRS-Bot-Nova~\cite{lhrsnova} pair large instruction corpora with
dedicated benchmarks, RS-LLaVA~\cite{rsllava} couples captioning with
VQA, VHM~\cite{vhm} adds explicit anti-hallucination training,
SkySenseGPT~\cite{skysensegpt} builds a 1.8M-sample instruction corpus on
a multiscale RS foundation model, and RS5M/GeoRSCLIP~\cite{georsclip}
scale vision-language pretraining for RS retrieval. These models
broaden \emph{what can be answered} from pixels, but answer every query
through the same monolithic generate-from-pixels interface: none decides
whether external capabilities are needed, nor routes to tool backends.
That is the axis our two-stage router addresses. \method{} is
complementary rather than competing: its Stage-B intrinsic path is
exactly such a VLM, and the trained router sits in front of it.

\paragraph{Agentic remote sensing and RL for tool use.}
RS-Agent~\cite{rsagent}, EarthAgent~\cite{earthagent} and
OpenEarthAgent~\cite{openearthagent} orchestrate expert models for earth
observation. Two OpenEarthAgent designs are adopted directly by our
Stage-B extrinsic path: \emph{two-turn trajectories with observation
masking} (tool-observation tokens are excluded from the training loss, so
the policy learns to \emph{use} observations rather than recite them) and
\emph{schema-registration-as-integration} (tools enter the catalog as
declarative schemas, no code changes).
\remoteagent{}~\cite{yao2026remoteagent} is closest to ours: it defines
the vague-intent routing problem, the \tcall{} action space and the
VagueEO protocol (10 task shapes, 5 unseen in training), reporting
95.0\% intent-recognition accuracy with
Qwen2.5-VL-7B~\cite{bai2025qwen25vl}, but its router consumes a downscaled
image and its GRPO stage covers intrinsic shapes only, leaving tool
routing a zero-shot capability. Vague queries as such are not unique to
remote sensing: ambiguous and underspecified visual questions have a
longer history in VQA, both as datasets that ground what an ambiguous
question could refer to~\cite{inadumi2024gazevqa} and as models that
learn to ask clarifying follow-ups~\cite{jian2025clearvqa};
\remoteagent{} is the work that carried this axis into EO agents, and
\vagueuhr{} extends it to UHR imagery with tool trajectories. Our
\vagueuhr{} corpus, explicit
tool-selection reward, and two-stage training recipe are direct
responses; we additionally provide both a TRL-based
path~\cite{vonwerra2020trl} and an ms-swift configuration mirroring the
paper's stack (\cref{sec:training}). GRPO~\cite{shao2024grpo}
group-normalizes sequence-level rewards and underpins verifiable-reward
(RLVR) training~\cite{guo2025r1}; \remoteagent{} applies answer-verifiable
rewards (Hungarian-matched IoU, relative-error decay, label coverage).
Also directly relevant to our cold-start strategy are ToRL's
tool-integrated RL (forced tool calls to seed
exploration)~\cite{li2025torl}, ToolRL's decomposition of tool rewards
into format and selection terms~\cite{qian2025toolrl}, and DAPO's
treatment of zero-variance groups~\cite{yu2025dapo}; together they
support ``SFT seeds the behavior, RL amplifies it'' pipelines
(\cref{sec:training,sec:whysft,sec:whyreward}).

\paragraph{Text-only routing and cascades.} Stage-A's design builds on a
line of text-only routing work. Toolformer~\cite{schick2023toolformer}
samples call sites from plain text corpora under self-supervision,
letting a model learn when to call, which API, and with which arguments,
evidence that tool-calling decisions are essentially text-decidable.
RouteLLM~\cite{ong2024routellm} trains routers that see only the query
text to dispatch between strong and weak models, roughly halving cost
without quality loss on several benchmarks and generalizing across
models, showing that query text carries sufficient routing signal.
Earlier cascading and query-complexity routing practice is consistent
with both. \method{} transfers this conclusion to
extrinsic/intrinsic discrimination and tool selection in UHR remote
sensing with a text-first decision interface (\cref{sec:router}); our
own $\pm$-image ablation then shows the transfer is only partial: the
trained policy's emission suppression is visually grounded
(\cref{sec:pmimage}).

\paragraph{Benchmarks and tool backends.} LRS-VQA~\cite{luo2025lrsvqa},
the remote-sensing subset of MME-RealWorld~\cite{li2024mmerealworld} and
XLRS-Bench~\cite{wang2025xlrs} are the standard UHR RS evaluation suites
(\cref{sec:setup} details our stratified-sampling protocol). For dense
prediction we rely on segmentation backends in the Segment Anything
family~\cite{kirillov2023sam,directsam2025}, exposed through MCP.

\paragraph{Differences from the closest systems.} Relative to
\weaveearth{}~\cite{ma2026weaveearth}: we add two-stage tool routing and
training; its evidence pipeline is adopted as Stage-B's optional
compression interface, with the recalibrated positioning of
\cref{sec:evidence}. Relative to \remoteagent{}~\cite{yao2026remoteagent}:
we add an explicit training signal for the routing decision stage
(alignment SFT $+$ $R_{\mathrm{WA2}}$), a two-turn tool-execution path
with an oracle attribution and repair of the conversion gap, a runnable
training stack, and a public corpus. Relative to zoom-and-search
methods~\cite{zoomeye,zoomsearch}: single-pass text-first routing
plus single-pass evidence construction replaces multi-round visual
exploration (efficiency comparison in \cref{sec:efficiency}).

\section{The \method{} Two-Stage Architecture}
\label{sec:method}

\method{} is implemented as the \texttt{weaveagent} package with six
subsystems (\texttt{evidence}, \texttt{skills}, \texttt{agents},
\texttt{mcp}, \texttt{training}, \texttt{data\_synthesis}), twelve
built-in tools behind a declarative MCP registry with automatic
discovery, and a publish/subscribe multi-agent layer. The
\texttt{skills} subsystem wraps every callable capability behind a
uniform interface (five built-in skill families covering evidence
retrieval, detection, segmentation, counting, and temporal comparison
front the
twelve built-in tools, and any MCP-registered tool can be exposed as a
skill without code changes); the \texttt{agents} subsystem provides
typed-event orchestration, and for all evaluations below the router runs
as one agent over this layer, keeping framework overhead out of the
reported latencies.

\subsection{Overview}
\label{sec:overview}

\begin{figure}[t]
  \centering
  \includegraphics[width=0.98\textwidth]{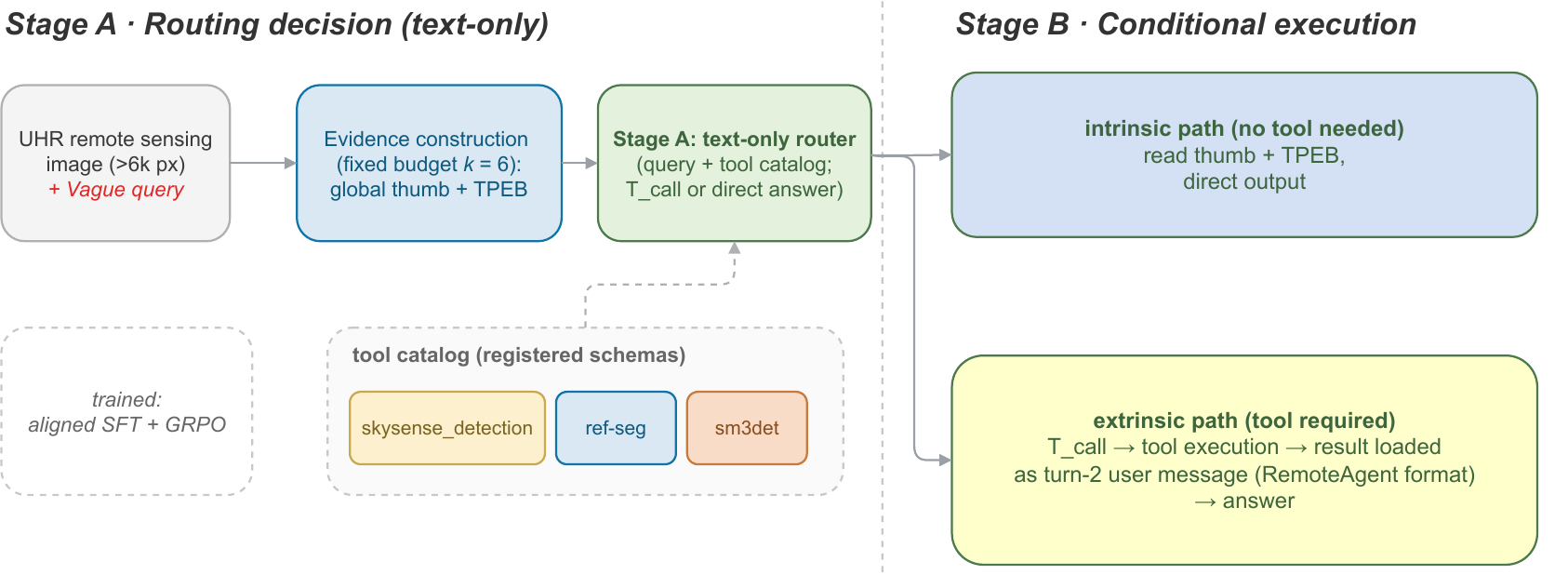}
  \caption{\method{} overview. Stage~A performs intrinsic/extrinsic
  discrimination and tool
  selection from the query and the catalog (a routing-first decision
  stage whose emission suppression the $\pm$-image ablation shows to be
  visually grounded; \cref{sec:router,sec:pmimage}). Stage~B executes
  conditionally: the intrinsic path answers from the thumbnail
  (optionally the \tpeb{}/\sem{} compression interface); the extrinsic
  path executes the \tcall{} on the \emph{original} full-resolution
  imagery and answers from the tool observation in a second round.}
  \label{fig:framework}
\end{figure}

\cref{fig:framework} gives an overview. Formally, given a query $q$ and a
tool catalog $C$ (text), Stage~A produces the routing decision
\begin{equation}
  a_{\mathrm{route}} \;=\; \pi_{\mathrm{route}}(q, C) \;=\;
  \begin{cases}
    \texttt{ANSWER}, & \text{intrinsic query: delegate to Stage-B visual
    answering},\\[2pt]
    \tcall(e_k, p), & \text{extrinsic query: tool $e_k$ applied to region
    $p$},
  \end{cases}
  \label{eq:route}
\end{equation}
and Stage~B executes conditionally on $a_{\mathrm{route}}$: the
intrinsic path generates \texttt{<answer>} spans from the visual input;
the extrinsic path executes the \tcall{}, backfills the tool observation
$o$ into the context, and generates the final \texttt{<answer>} in a
second round. The whole system is driven by the same backbone
(Qwen3-VL-2B~\cite{bai2025qwen3vl}) trained with the recipe of
\cref{sec:training}; Stage-A's \emph{decision interface} is designed to
require only text (query $+$ catalog). In the unified prompt used for
all reported training and evaluation runs, the thumbnail and evidence
board are also present (matching the answering-path format), so that
trained behavior transfers unchanged to evaluation; whether the routing
decision itself depends on those visual tokens is exactly what the
$\pm$-image ablation isolates. It does: emission suppression is
visually grounded (\cref{sec:pmimage}).

\paragraph{Why decouple.} (i) \emph{Token economics}: the decision stage
is text-first by design; UHR visual tokens should be spent only after the
decision, and only on the selected path (how far the trained policy
attains the text-only ideal is measured in \cref{sec:pmimage}). (ii)
\emph{Format reliability}:
the \tcall{} is a textual decision that alignment training can make
reliably parseable (\cref{sec:whysft}); a decision interface mixed with
visual tokens inherits both visual and format fragility. (iii) \emph{A
loss-free tool path}: tools act on the original full-resolution imagery,
so extrinsic queries are never bounded by visual compression; the
compression interface serves the intrinsic path only
(\cref{sec:intrinsic}).

\subsection{Stage A: Text-Only Router}
\label{sec:router}

\paragraph{Input and output.} Stage~A reads two pieces of text: the query
$q$ and the tool catalog $C$, the latter rendered automatically from the
declarative MCP registry~\cite{anthropic2024mcp} (tool names, functional
descriptions, argument schemas). Its output is the routing decision of
Eq.~\eqref{eq:route}: intrinsic queries are delegated to answering;
extrinsic queries select a tool $e_k$ and emit $\tcall(e_k, p)$, with $p$
grounded in the catalog's region syntax. Final answering after the tool
observation returns belongs to Stage~B (\cref{sec:extrinsic}).

\paragraph{Design basis.} Tool-routing decisions are largely determined
by the query text: Toolformer~\cite{schick2023toolformer} shows that
pure-text self-supervision suffices to learn the three elements of tool
calling (when, what, with which arguments);
RouteLLM~\cite{ong2024routellm} shows that text-only routing halves cost
without quality loss and generalizes across models. Stage-A's decision
interface is therefore designed to need no visual tokens: this would decouple
routing cost from image size entirely (routing a 27{,}328-px image costs
the same as routing a 4{,}096-px one), and would reduce ``routing
generalization to unseen tools'' to a pure-text problem of adding one
schema entry to the catalog. The sm3det measurement of \cref{sec:unseen}
shows registration alone does not deliver this ($0/62$ on the unseen
tool), so this remains design intent, not a demonstrated property. In the unified
prompt of \cref{sec:overview} the visual inputs are nonetheless present.

Our own $\pm$-image ablation (E19; \cref{sec:pmimage}) shows that the
strong form of this design basis does \emph{not} hold for the trained 2B
policy: with image tokens removed (identical prompts and records),
extrinsic routing is preserved ($0.8375$ vs.\ $0.8075$) but intrinsic
routing collapses ($0.985\!\to\!0.0217$) and the model emits a call on
almost every record (987 of 1{,}000). The visual input is the signal
that suppresses emission on intrinsic queries: knowing \emph{when not to
call} is visually grounded in UHR imagery. The $15\times$ latency gap
($0.257$\,s vs.\ $3.85$\,s per record) makes a truly text-only router an
attractive but unattained target; distilling one is future work
(\cref{sec:limitations}).

\paragraph{Action space.} Following \remoteagent{}, extrinsic decisions
emit the hybrid action $\tcall(e_k, p)$ with $e_k$ chosen from the
injected catalog. The behavior is strictly binary (emit a \tcall{} or
delegate to answering), with no degenerate output forms (smoke audit,
E12). The direct ablation of zero-visual-token vs.\ image-conditioned
routing has been run (E19, \cref{sec:pmimage}): removing image tokens
keeps extrinsic routing but collapses intrinsic routing and re-opens
emission on almost every record, so Stage~A rests on the literature
above, on the end-to-end routing measurements of \cref{sec:routing}, and
on this measured boundary of the text-only design basis.

\subsection{Stage B (Intrinsic Path): Visual Answering with an Optional
Compression Interface}
\label{sec:intrinsic}
\label{sec:evidence}

\paragraph{Default path.} The answering input for an intrinsic query is
the full-scene 2048-px thumbnail plus the query text, exactly the
observation interface of the static VQA baseline.

\paragraph{Optional compression interface (\tpeb{} $+$ \sem{}).} For
fine-grained task types that need local detail, \method{} offers an
optional structured compression interface that compresses an
arbitrary-size UHR image into a \emph{fixed token budget} (thumbnail
$+$ evidence board, two views ${\approx}5$k tokens). Evidence
construction adopts the \weaveearth{} pipeline; our configuration is
aligned with the authors' released script item by item (encoder
\texttt{siglip2-so400m-patch16-naflex}~\cite{zhai2025siglip2}, grid
$6\times6$ with overlap $0.1$, anchor top-8, evidence budget $k{=}6$,
448-px evidence patches). Given image $I$ and query $q$, candidate
patches $\{p_i\}$ are sampled on the overlapping grid and encoded jointly
with the query and the global thumbnail $g$; each candidate receives a
Global Context Constraint (\gcc{}) score
\begin{equation}
  s_i \;=\; \mathrm{sim}(p_i, q) \;+\; \lambda\,\mathrm{sim}(p_i, g),
  \qquad \lambda = 0.35,
  \label{eq:gcc}
\end{equation}
where both similarity terms are cosine similarities in the SigLIP2 embedding
space. The top-$a$ anchors ($a{=}8$) feed a training-free greedy pass that
selects the \mses{} under the budget $k$:
\begin{equation}
  S^{*} \;=\; \arg\max_{S \subseteq C,\ |S| \le k}\;
  \mathrm{Rel}(S,q) \;+\; \alpha\,\mathrm{Cov}(S)
  \;-\; \beta\,\mathrm{Red}(S),
  \label{eq:mses}
\end{equation}
trading off relevance, coverage and redundancy. Selected regions are expanded
with their grid neighbors to avoid fragmentation and woven into the
\tpeb{}: a compact board in which each selected region occupies a cell
while the \emph{relative spatial topology} of the scene is preserved.
Every board unit is described by a \sem{} tuple
$e_i = (p_i, u_i, b_i, n_i, r_i, \ell_i)$: patch identifier, role
(anchor/support), grid position, bounding box, neighbor set, and scale,
rendered into the prompt, e.g.\
\begin{quote}\small
\texttt{[Patch R14 | role=support | grid=(2,1) | box=(0.250,0.417,0.200,0.200) |}\\
\texttt{~~~~neighbors=\{R7,R8,R9,R13,R15,R19,R20,R21\} | scale=grid\_6x6]}
\end{quote}
Coordinate semantics are $(cx, cy, w, h)$ normalized bounding boxes;
evidence patches are resized, never upsampled.

\paragraph{Positioning: compression interface, not booster.} The benefit
is \emph{token determinism}, not accuracy: token cost is fixed
(two views, ${\approx}5$k tokens) regardless of source size, whereas the
naive detail-preserving alternative (high-resolution crop stitching) has
a token cost that grows linearly with source size and detail demand, and
the thumbnail baseline, though equally fixed-cost, abandons local detail
entirely. The costs are just as concrete: a $1.31\times$ detail
ceiling (\cref{sec:ceiling}), coverage of only 6 of 36 grid cells, and
destroyed global spatial continuity; on three static VQA benchmarks the
paired deltas against the thumbnail baseline are $-1.0/-3.4/-2.9$ points,
and component ablations show no positive contribution
(\cref{sec:ablation}). We therefore position it as an
\emph{optional compression interface for fine-grained task types},
enabled per task type, and make no accuracy-improvement claim.

\paragraph{Implementation faithfulness.} We reimplement the \mses{}
selection (Eq.~\eqref{eq:mses}) \emph{per the paper's algorithm}: in the
authors' released code the candidate pool is truncated to the top-8
anchors before the budget-$k$ greedy loop, so the loop never executes (a
neutral implementation-difference record, \cref{sec:repro}); the
paper-faithful reimplementation is therefore retained rather than aligned
to the release. Shared components were audited item by item and agree:
encoder and scoring formula, grid/overlap/anchor/board parameters,
coordinate semantics, board labels (byte-identical), resize behavior, and
answer-matching rules; the $k$ pipeline was verified with a budget probe
($k{=}2/4/6/8/10$ yield $2/4/6/8/10$ regions).

\subsection{Stage B (Extrinsic Path): Tool Execution, Observation
Backfill, and Two-Turn Trajectories}
\label{sec:extrinsic}

\paragraph{Execution.} After Stage~A emits $\tcall(e_k, p)$, the system
executes the tool on the \emph{original full-resolution image}: the
extrinsic path passes through no visual compression, so detection,
segmentation and counting precision are bounded only by the tool backend
itself. Backends are pluggable with a three-tier fallback (native MCP
stdio servers $\rightarrow$ released HTTP expert services $\rightarrow$
generic JSON endpoints), so the same router drives both real deployments
(e.g., DirectSAM-class segmentation~\cite{directsam2025}) and
annotation-grounded simulated backends; we disclose per experiment which
tools are real.

\paragraph{Observation backfill and two-turn trajectories.} Tool outputs
(detection boxes, mask statistics, counts) are serialized into a
structured observation $o$ and backfilled into the context; the second
round generates the final \texttt{<answer>} from $(q, C, o)$. Training
follows OpenEarthAgent's~\cite{openearthagent} \emph{two-turn
trajectories with observation masking}: observation tokens are excluded
from the loss, so the policy learns to \emph{use} observations rather
than recite them.

\paragraph{Schema registration as integration.} Because tools register by
declarative schema and the catalog is text (\cref{sec:router}), an
unseen tool can be injected into the catalog after training to test
routing generalization without retraining (\cref{sec:unseen}).

\paragraph{Known gap, diagnosed and partially repaired.} With extrinsic
routing at 80.75\%, final-answer accuracy on extrinsic tasks remained low
(detection $0.035$--$0.045$, segmentation $0.005$ across E13/E15 under
the single-turn protocol): an observation$\to$answer \emph{conversion
gap}. An oracle
attribution study (\cref{sec:conversion}) localizes the break to the
conversion of marker-free tool text rather than to calling quality, and a
two-turn trajectory SFT stage (\cref{sec:twoturn}, evaluated in
\cref{sec:conversion}) partially repairs it under protocol caveats
disclosed there.
\subsection{Training Recipe: Alignment SFT $\rightarrow$ $R_{\mathrm{WA2}}$
GRPO}
\label{sec:training}

Training has two steps, each addressing a failure mode that must be fixed
separately (experimental rationale in \cref{sec:whysft,sec:whyreward}).

\paragraph{Step 1: alignment SFT (seeding $+$ format alignment).} The
cold-start set is built from 870 ground-truth tool-trajectory
demonstrations (syntax character-identical to the parser, arguments with
all required keys; histogram ref-seg $579$ / skysense-det $291$) plus 870
intrinsic balancing demonstrations (to prevent indiscriminate calling),
\emph{rebuilt with prompts and image formats strictly identical to
evaluation}: training and evaluation prompts are byte-identical at the
non-image level, and the difference list of Table~S1 is,
configuration by configuration, exactly the fix list). SFT configuration: Qwen3-VL-2B~\cite{bai2025qwen3vl},
LoRA $r{=}32/\alpha{=}64$~\cite{hu2022lora} on all linear layers, 1{,}000
steps, effective batch 16, lr $10^{-5}$ cosine with warmup; loss
$1.73\to0.217$, 273.9\,min, peak 14.8\,GB. The acceptance gate is an
\emph{evaluation-path smoke test} (threshold $0.80$; passed).

\paragraph{Step 2: GRPO with the routing-first reward.} We compare
sequence-level rewards under GRPO~\cite{shao2024grpo} with
group-normalized advantages
$\hat A_i = (r_i - \bar r)\,/\,\mathrm{std}(r)$ over $G$ rollouts per
prompt: $G{=}4$ generations, rollout temperature $0.95$, learning rate
$10^{-6}$, 300 steps ($\approx$1.4 prompt-epochs over the 1{,}740 aligned
rows), generation batch 8 (effective
optimization batch 32 under TRL 1.12.0; the small generation batch cuts
generation memory to a quarter~\cite{vonwerra2020trl,kwon2023vllm}),
fitting on a single 24-GB GPU (measured peak 13.8--15.8\,GB). Training
rows carry thumbnail paths (on disk, lazily decoded) and the tool-catalog
text; the \sem{} block is disabled during training to match the
evaluation-time observation interface. LoRA weights are merged before
evaluation. \remoteagent{}'s official stack is ms-swift with DeepSpeed
ZeRO-2 and trains only on intrinsic tasks; we provide both a TRL-based
path~\cite{vonwerra2020trl} and an ms-swift configuration mirroring the
paper's stack in the released code. The 8B backbone could not be
GRPO-trained within 24\,GB (\cref{sec:limitations}).

\paragraph{Rewards.} The paper-faithful answer-only reward
$R_{\mathrm{RA}}$~\cite{yao2026remoteagent} scores only the
\texttt{<answer>} field, dispatched by ground-truth format:
\begin{equation}
  R_{\mathrm{RA}}(a_{\mathrm{pred}}, a_{\mathrm{gt}}) =
  \begin{cases}
    R_{\mathrm{coord}}, & \text{coordinate gt (Hungarian-matched IoU)},\\
    R_{\mathrm{num}},   & \text{scalar gt }(\,e^{-3\,|p-g|/|g|}\text{ with
    edge-case rules)},\\
    R_{\mathrm{text}},  & \text{label gt (coverage } |gt \cap pred| /
    |gt|\,\text{)},
  \end{cases}
  \label{eq:reward-ra}
\end{equation}
matching \remoteagent{}'s published reward. Our
\emph{UnifiedMultimodalReward} $R_{\mathrm{WA}}$ adds two explicit terms:
\begin{equation}
  R_{\mathrm{WA}} \;=\; R_{\mathrm{ans}}
      \;+\; w_t\,\mathbb{1}[\hat e = e^{*}]
      \;+\; w_f\,\mathbb{1}[\text{well-formed}],
  \label{eq:reward-wa}
\end{equation}
where $R_{\mathrm{ans}}$ is a format-aware answer match (normalized
exact / integer / substring / token-F1 $\ge 0.8$ / MCQ letter match),
$\hat e$ is the chosen tool, $e^{*}$ the trajectory tool; the measured
configuration is $w_a{=}1.0$, $w_t{=}0.3$, $w_f{=}0.1$, plus a $0.2$
consolation score for extrinsic no-calls.
The per-case scores are listed in Appendix~B
(Table~S4). The tool-selection
term is the key design divergence (\remoteagent{} never trains tool
routing, whereas Eq.~\eqref{eq:reward-wa} rewards \emph{choosing the
right tool}), but the structure has three economic defects: (i) the
SFT-taught behavior (a bare \tcall{}) caps at $0.35$, only $0.24$ above
``no-call $+$ wrong answer'' ($0.11$); (ii) answering an extrinsic task
correctly ($1.11$) is worth $3.2\times$ a correct call ($0.35$); (iii)
intrinsic tasks are 73\% of the routing-training corpus (2{,}400/3{,}273), and calling
on them scores $0$. The aggregate consequence
(\cref{tab:rewardexpect}): with $p_{\mathrm{ans}}$ the policy's answer
accuracy on extrinsic tasks, the expected reward of always answering
directly ($0.11 + p_{\mathrm{ans}}$) exceeds that of correct routing
($0.35$) once $p_{\mathrm{ans}} \ge 0.24$: the RL-rational policy is
``answer everything''. E12's measured extrinsic direct-answer accuracy
is $p_{\mathrm{ans}} = 0.285$ (it answers directly on 400/400 extrinsic
rows), just above the 0.24 indifference threshold. ``Always answer
directly'' was indeed the empirically rational policy for the unaligned
model, and $R_{\mathrm{WA2}}$'s incentive inversion is designed to move
the policy away from that equilibrium.

\paragraph{The routing-first revision $R_{\mathrm{WA2}}$.} The revision
is
\begin{equation}
  R_{\mathrm{WA2}}(\text{response}) \;=\; 1.0\, r_{\mathrm{ans}}
  \;+\; 1.0\, r_{\mathrm{tool}} \;+\; 0.5\, r_{\mathrm{fmt}},
  \label{eq:reward-wa2}
\end{equation}
with indicator rules transcribed from the released
\texttt{reward\_wa2.py} ($R_{\mathrm{WA}}$ keeps the $1.0/0.3/0.1$
weights of Eq.~\eqref{eq:reward-wa} and the format rules below):
$r_{\mathrm{ans}} = 1$ iff the \texttt{<answer>} span matches the ground
truth, and on extrinsic tasks a \emph{correct route is credited with the
answer term automatically} ($r_{\mathrm{ans}} = \max(r_{\mathrm{ans}},
1)$, since the answer comes from the tool); $r_{\mathrm{tool}} = 1$ iff the
emitted call matches the trajectory tool, and $0$ for a wrong call or a
\emph{silent skip}: the $0.2$ no-call consolation of $R_{\mathrm{WA}}$
is deleted (silent skip $= 0$); on intrinsic tasks $r_{\mathrm{tool}} =
1$ for correctly not calling and $0$ for a stray \tcall{};
the format term contributes up to $1.0$ unweighted ($0.5$ per
well-formed component: the answer span and the \tcall{}), weighted by
$w_{\mathrm{fmt}}=0.5$; the canonical extrinsic emission, a bare,
well-formed \tcall{} without an answer span, earns $0.25$, giving
correct routing $2.25 = 1.0 + 1.0 + 0.25$ (\cref{tab:rewardexpect}).
Three structural modifications follow:
\emph{delete the no-call consolation score ($0.2\to0$), raise the
routing weight ($0.3\to1.0$), and split the format term into two $0.5$
components}. Correct direct answers actually rise ($1.11\to1.25$),
but correct routed calls rise more ($0.35\to2.25$), so the
expectations
invert
(\cref{tab:rewardexpect}): correct routing $2.25$ $\gg$ direct answering
$0.25$--$1.25$ per case, and the inversion is parameter-free; it holds for any
answer accuracy. Under $R_{\mathrm{WA}}$ the
RL-rational policy is all-direct-answer (measured: mean completion
length converges to 16--25 tokens, i.e.\ no calls); under
$R_{\mathrm{WA2}}$ it is routing (training dynamics healthy,
\cref{tab:dynamics}).

\begin{table}[t]
\centering\small
\caption{Per-case rewards on extrinsic tasks and the expectation algebra
(\texttt{reward\_wa2.py}; per-case values, no measured capability
parameters). Weights: $R_{\mathrm{WA}}$ $=$ $1.0$ answer $+$ $0.3$ tool
$+$ $0.1$ format with a $0.2$ no-call consolation; $R_{\mathrm{WA2}}$
$=$ $1.0$ answer $+$ $1.0$ tool $+$ $0.5$ format, no consolation, and a
correct route is automatically credited with the answer term
($r_{\mathrm{ans}} = \max(r_{\mathrm{ans}}, 1)$;
Eq.~\eqref{eq:reward-wa2}). With $p_{\mathrm{ans}}$ the policy's answer
accuracy on extrinsic tasks,
$E_{R_{\mathrm{WA}}}[\text{always-answer}] = 0.11 + p_{\mathrm{ans}}$
exceeds $E[\text{correct routing}] = 0.35$ once $p_{\mathrm{ans}} \ge
0.24$: all-direct-answer is RL-rational. Under $R_{\mathrm{WA2}}$ the
inversion ($2.25 \gg 0.25 + p_{\mathrm{ans}}$) is parameter-free; it
holds for any $p_{\mathrm{ans}} \le 1$.}
\label{tab:rewardexpect}
\begin{tabular}{lcc}
\toprule
Policy (on extrinsic tasks) & $R_{\mathrm{WA}}$ & $R_{\mathrm{WA2}}$ \\
\midrule
Correct routing (right tool $+$ call) & 0.35 & \textbf{2.25} \\
Wrong tool call & 0.05 & 0.25 \\
Silent skip (no call, no answer) & 0.06 & \textbf{0.0} \\
Always answer directly: correct\,/\,wrong & 1.11\,/\,0.11 & 1.25\,/\,0.25 \\
\bottomrule
\end{tabular}
\end{table}

\subsection{Two-Turn Trajectory SFT: Repairing the
Observation$\to$Answer Conversion}
\label{sec:twoturn}

\begin{figure}[t]
  \centering
  \includegraphics[width=0.98\textwidth]{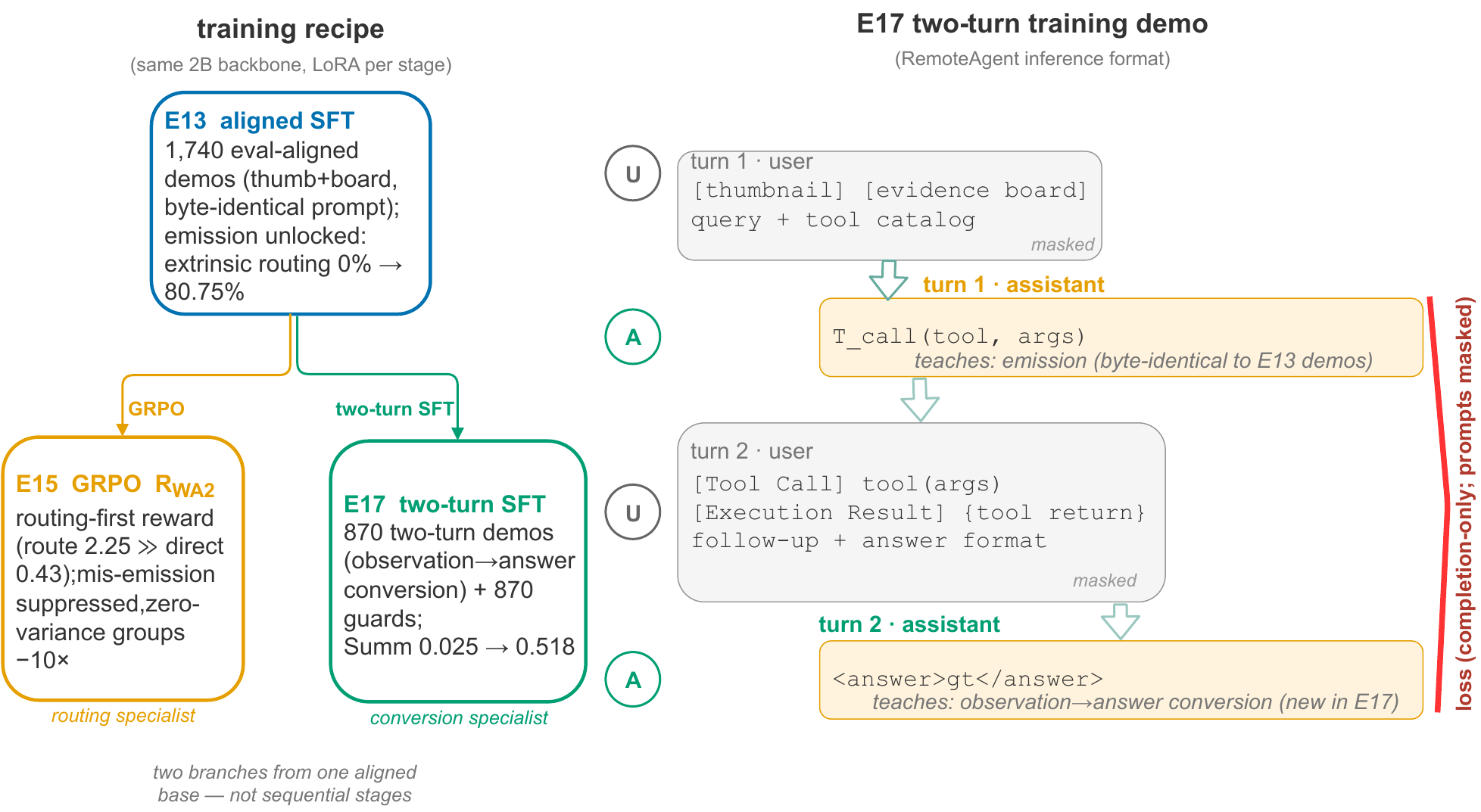}
  \caption{The training recipe (left): alignment SFT and the E15 GRPO
  stage on the single-turn branch; right: the E17 two-turn stage, which
  branches from the E13 base. Turn~1 is byte-identical to the E13
  aligned demonstrations; turn~2 loads the tool observation in the
  \remoteagent{} inference format. The completion-only loss covers both
assistant turns: emission (inherited) and observation$\to$answer
conversion (new in E17).}
  \label{fig:training}
\end{figure}

The oracle study of \cref{sec:conversion} motivated a third training
stage on top of the alignment-SFT merged weights (the E13 base),
as illustrated in \cref{fig:training}. The
fine-tuning uses 1{,}740 rows: 870 \emph{two-turn tool demonstrations}
(turn~1 identical to the E13 prompt, image and \tcall{}; turn~2 a
\texttt{[Tool Call]} $+$ \texttt{[Execution Result]} exchange followed by
the \remoteagent{}-style follow-up prompt, with the observation stuffed
in the embedded format of \cref{sec:conversion}), plus 870 single-turn
intrinsic demonstrations (anti-regression). Configuration: LoRA
$r{=}32/\alpha{=}64$ on q/k/v/o with dropout $0.05$, lr $10^{-5}$ cosine
with 3\% warmup, batch $1\times$ accumulation 16, 1{,}000 steps, bf16
with gradient checkpointing, max length 8{,}192; 276\,min total, final
loss $0.165$, peak 15.1\,GB. The stuffing format replicates, verbatim,
the \texttt{[Tool Call]}/\texttt{[Execution Result]} loop of
\remoteagent{}'s released inference script. All returns used in training
are oracle/annotation-grounded; the evaluation protocols are disclosed
in \cref{sec:conversion}. E17 therefore shares the E13 base rather than
inheriting the E15 GRPO checkpoint: E15 and E17 are two branches from
E13 (a single-turn GRPO specialist and a two-turn SFT specialist), not
sequential stages, and E15--E17 differences reflect the branch rather
than a third stage.

\section{The VagueUHR Corpus}
\label{sec:data}

No public corpus links vague UHR intents to tool-call trajectories:
VagueEO~\cite{yao2026remoteagent} ships an evaluation protocol but trains
on intrinsic tasks only, and \weaveearth{}-style benchmarks contain
answers but no tool traces. \vagueuhr{} fills the gap
\emph{synthetically}: trajectories are constructed on top of established
UHR benchmarks so that every answer is derived from existing annotations
and never model-generated.

\subsection{Synthesis Pipeline}
\label{sec:synthesis}

Records are bootstrapped from LRS-VQA, MME-RealWorld-RS and XLRS-Bench
imagery and annotations. The pipeline has three stages: (i) a
\emph{query generator} maps each source record to one of ten task shapes
and emits a deliberately vague query via type-preserving vagueness
operators from a template bank; (ii) a \emph{trajectory synthesizer}
constructs the reference plan (an intrinsic answer, or
$\tcall(\text{tool}, \text{region})$ steps grounded in the annotations)
together with the assumed tool catalog; (iii) a rule-based
\emph{validator} checks query--annotation--trajectory consistency,
schema compliance, and jargon leakage, rejecting any failing record; the
shipped corpus passes validation at 100\%. The LLM query-paraphrase
channel is \emph{not} enabled on the training side;
the test-side LLM-visual-rewrite split is described next.

\subsection{Dual Splits and Statistics}
\label{sec:splits}

\cref{tab:data} summarizes the corpus. One documented deviation from the
VagueEO protocol: absent bi-temporal imagery in the sources, the
change-detection shape collapses to detection. The 873 tool-trajectory
records in the routing split teach the \emph{syntax and mechanics} of
\tcall{} over the training-visible tool categories; five of the ten
evaluation shapes never appear in training, so routing \emph{to those
shapes} measures generalization, not memorization.

\begin{table}[t]
\centering\footnotesize\setlength{\tabcolsep}{4pt}
\caption{\vagueuhr{} dual splits. Tool-trajectory records carry explicit
$\tcall$ steps; all answers derive from source annotations. Five of the
ten evaluation task shapes never appear in training; the test tool pool
contains one training-unseen tool (sm3det, 62 records).}
\label{tab:data}
\begin{tabular}{p{2.3cm}p{2.2cm}p{3.6cm}p{3.4cm}p{3.4cm}}
\toprule
Split & Records & Tool-traj.\ records & Tool pool & Task shapes \\
\midrule
Train (template) & 5{,}000 & --- (all intrinsic) & --- & 5
(training-visible) \\
Routing-train (\texttt{train\_tools}) & 3{,}273 & 873 (870 rendered
tool-call demos; 3 skipped: gt trajectory has no real tool step; 0
render failures) & 2 visible: ref-seg 579 / skysense-det 291 & 2{,}400
intrinsic $+$ 873 tool \\
Test (template) & 1{,}000 & 400 & 3: skysense 138 $+$ sm3det 62
(unseen) $+$ ref-seg 200 & 10 (5 unseen in training); 600 intrinsic
$+$ 400 tool \\
Test (LLM-rewritten) & 992 kept $+$ 8 removed (gt--image conflicts)
$=$ 1{,}000 & 593 intrinsic $+$ 399 extrinsic (8 removed) & same & same \\
\bottomrule
\end{tabular}
\end{table}

\paragraph{Role of each split.} The 5{,}000-record template split is the
\emph{synthesis base}: the 3{,}273-record routing-training split (with
tool trajectories), both test splits, and the clear-register control of
\cref{sec:register} are all derived from it. It is not itself used for
optimization: every optimization stage consumes a 1{,}740-row aligned
set drawn from the 3{,}273 routing-training records (870 tool
demonstrations with trajectories $+$ 870 intrinsic guards): the
cold-start and alignment SFT stages train on it directly, the GRPO
stage optimizes on it, and the two-turn stage converts its tool half
into two-turn trajectories while reusing the guards. A
clear-register control split (\texttt{VagueUHR\_test\_clear.jsonl},
1{,}000 records) is additionally built for \cref{sec:register}: the
original benchmark questions are recovered by joining on image id and
answer (276 ambiguous keys resolved by token overlap with the vague
query; the instruction tail stripped), replacing only
\texttt{query.text}.

\paragraph{LLM-visually-rewritten test split.} An LLM (GLM-5.3-Flash)
rewrote the template queries \emph{per image, looking at the image}:
992 records kept, 8 skipped for ground-truth--image conflicts; 100\%
unique queries, 143 sentence-initial word types, zero template
phrasing; ground truth, images and task types were audited
independently with \emph{character-level zero drift}. Diversity
contrast (template $\to$ rewritten): distinct queries 92.6\% $\to$
100\%; ``What\ldots'' openings 35.7\% $\to$ 0.9\%; sentence-initial
word types 31 $\to$ 143. The two test splits coexist so that query
surface wording can be treated as a controlled variable; the paired
wording-sensitivity measurement is \cref{sec:wording}.

\paragraph{Clear-register split.} A third, clear-register variant is
shipped in the corpus directory \texttt{datars/}
(\texttt{VagueUHR\_test\_clear.jsonl}) and serves as the
clear column of the register matrix (\cref{tab:register}): unambiguous queries
are reconstructed by joining each test record to its source-benchmark
question via image id $+$ answer, token-overlap disambiguation resolves
all 276 ambiguous joins, and instruction tails are stripped.

\subsection{Unseen-Tool and Unseen-Shape Design}
\label{sec:unseen}

The 400 test tool tasks use three tools:
\texttt{skysense\_detection} (detection/counting; 138 records),
\texttt{referring\_expression\_}\linebreak\texttt{segmentation}
(segmentation; 200
records), and \texttt{sm3det\_oriented\_detection} (62 records), the
last \emph{entirely unseen in training}. Because the catalog is text and
registration is declarative (\cref{sec:router}, \cref{sec:extrinsic}),
routing to sm3det tests generalization to a schema entry never seen in
training. The result is negative:
registration makes the tool \emph{visible} (its schema enters the
catalog) but not \emph{usable}. On the 62 unseen-tool queries all
three trained checkpoints (E13/E15/E17) route $0/62$, with emissions
collapsing onto the seen tools (\texttt{skysense\_detection} 48,
\texttt{referring\_expression\_segmentation} 3, no call 11 for E15). Tool
generalization requires exposure, not registration. Only two tools are
training-visible versus three at test time,
an intentional asymmetry. Other catalog tools (e.g.,
\texttt{change3d\_bcd}, \texttt{region\_contour\_extraction}) remain
available to the router at inference. The same logic applies to task
shapes: five of the ten evaluation shapes never appear in training. The
tool-selection metrics of \cref{sec:routing} are computed over all 400
extrinsic records, including the 62 unseen-tool records.

\subsection{Annotation Audit}
\label{sec:honesty}

Four disclosures: (i) all template-split records carry
\texttt{vague\_level} $= 0.5$ and \texttt{source} $=$ heuristic:
vagueness is templated and narrower than open-ended user phrasing; (ii)
\emph{11\% of test answers are unseen in training}; (iii) 34 loose-count
queries carry hedge wording; (iv) the six test records drawn from the
two 1213 source images (the cluster flagged by the annotation audit)
were manually re-checked against the imagery: five labels were
confirmed, and one was identified as a source-annotation misread (the
``surrounding area'' query labeled \emph{urban} whose surroundings are
rural); the record is flagged for correction in the next corpus
revision and is retained here unchanged so that the test set remains
byte-identical to the released version. The 8 removals in the rewritten
split are precisely the output of the annotation--image consistency
check.

\section{Experiments}
\label{sec:expts}

\subsection{Setup and Stratified-Sampling Protocol}
\label{sec:setup}

\paragraph{Benchmarks and stratified sampling.} We evaluate on LRS-VQA
(7{,}333 questions; 8 categories), the remote-sensing subset of
MME-RealWorld (3{,}738 multiple-choice questions; Color, Count, Position),
and XLRS-Bench-lite (3{,}080). Because full-benchmark sweeps are
prohibitive on our single-GPU budget, all our numbers are computed on
category-proportional stratified samples: 1{,}500 records per benchmark
for the compression-interface characterization (seed 0; all 8 LRS
categories covered with proportions matching the full set; MME position
33.6\% / color 33.5\% / count 32.9\%), 500 per variant for the component
ablations and the $k$-scan, and 300 for the official-script comparison.
Stratified estimates are unbiased but carry sampling noise; we therefore
report \emph{paired} evidence-on/off comparisons on identical samples
and, where possible, McNemar tests. \emph{Our stratified estimates are
not directly comparable to paper-reported full-benchmark numbers}, and we
never place them in the same row without saying so. On XLRS we can run 7
of the 8 subcategories (SR is unavailable to us; OP contributes 53.9\% of
our sample), so per-subcategory columns cannot be matched to the
\weaveearth{} paper's Table~2; overall (weighted) accuracy remains
comparable.

\paragraph{Backbone-to-claim map.} Two backbones appear, each tied to
specific claims: the \textbf{trained 2B} (Qwen3-VL-2B, all of
E13/E15/E17) carries every routing, reward, and conversion result
(\cref{sec:routing,sec:conversion}); the \textbf{zero-shot 8B}
(Qwen3-VL-8B, E3a and the E14 wording arms) is used only as the
zero-shot floor and its wording-robustness check, plus the compression-
and efficiency-interface characterizations
(\cref{sec:ablation,sec:efficiency}), which involve no training. No
table mixes backbones within a comparison; the 2B scale limitation is
\cref{sec:limitations} item~1.

\paragraph{Decoding and matching.} All evaluations use \emph{greedy
decoding} (temperature 0), matching the official scripts: paired
comparisons are the measurements most sensitive to decoding noise, and a
$\pm1$--2-point sampling-noise floor swamps most claimed gains at these
sample sizes. Our earlier sampling-mode runs were an accident; those
outputs are quarantined (the quarantine boundary is every
sampling-mode output directory), and no number in this paper comes from
them.
Answer matching follows \weaveearth{}'s released rules (normalized exact
/ integer / substring / token-F1 $\ge 0.8$) plus MCQ letter matching;
MCQ option instructions must be ordered \emph{after} the \sem{} region
metadata (\cref{sec:repro}).

\paragraph{Backbones and hardware.} Compression-interface and
intrinsic-path evaluations use Qwen3-VL-8B (matching \weaveearth{})
served by vLLM 0.27.1~\cite{kwon2023vllm} (FP8) on a single RTX 4090
(24\,GB); routing training and all two-stage evaluations use
Qwen3-VL-2B. \remoteagent{}'s official numbers use Qwen2.5-VL-7B under
its own protocol and are reported separately; we do not co-scale them
with ours.

\paragraph{Positioning with respect to public benchmarks.} All headline
numbers are measured on \vagueuhr{}-test rather than on public RS
suites. This is deliberate: the quantities under study (whether to
call, what to call, and whether a call converts into an executed
observation) are properties of the query--tool interaction protocol,
not of single-image perception, and public suites expose no executable
tool backends, so none of our headline metrics is reproducible on them.
Cross-model perception rankings on those
suites~\cite{geochat,earthgpt,rsgpt,lhrsnova} answer a different
question and are treated as related work, not baselines; the
tool-execution ceiling is instead bounded from above by the oracle arms
(\cref{sec:conversion}).

\subsection{Routing and Tool Metrics on \vagueuhr{}-test}
\label{sec:routing}

\begin{table}[t]
\centering\footnotesize\setlength{\tabcolsep}{4pt}
\caption{Routing and tool metrics on \vagueuhr{}-test (1{,}000 records;
600 intrinsic $+$ 400 tool-requiring; five task shapes unseen in
training; all greedy). Intr.\ = intrinsic routing accuracy; Extr.\ =
extrinsic routing accuracy; T-sel.\ = fraction of \emph{all} records
whose tool decision is correct (a correct no-call on an intrinsic
record counts as correct, as does an emitted call matching the
ground-truth tool on an extrinsic record; hence E13's
$0.905=(0.970{\times}600+323)/1000$); Calls = number of \tcall{}
emissions; Mean = overall
answer accuracy over all 1{,}000 records. Call accounting: E13's 398
$=$ 380 extrinsic single-calls $+$ 18 intrinsic mis-emissions; E15's
384 $=$ 375 $+$ 9. The native-channel row emits through the official
\texttt{tools=} channel rather than \tcall{}; its Mean is not measured.
$^{*}$With zero emissions the
tool-selection aggregate is a bookkeeping artifact (nothing to judge),
not a measured capability.}
\label{tab:routing}
\begin{tabular}{lccccc}
\toprule
Model / training stage & Intr. & Extr. & T-sel. & Calls & Mean \\
\midrule
Qwen3-VL-8B zero-shot (E3a) & 1.000 (600/600) & 0.000 (0/400) &
0.600$^{*}$ & 0 & 0.2630 \\
GRPO $R_{\mathrm{RA}}$ from 2B base (E3b) & 1.000 & 0.000 & 0.600$^{*}$
& 0 & 0.2420 \\
GRPO $R_{\mathrm{WA}}$ from 2B base (E3b) & 1.000 & 0.000 & 0.600$^{*}$
& 0 & 0.2570 \\
\midrule
Base 2B, native tools= channel (zero-shot) & 0.015 (9/600) & 0.12
(48/400) & 0.057 & 986 & --- \\
\midrule
Aligned SFT (E13, 2B) & 0.970 & \textbf{0.8075} (323/400) & 0.905 & 398
& 0.2430 \\
\quad $+$ GRPO $R_{\mathrm{WA2}}$ (E15) & \textbf{0.985} &
\textbf{0.8075} (323/400) & \textbf{0.914} & 384 & \textbf{0.2500} \\
\bottomrule
\end{tabular}
\end{table}

\paragraph{The pre-training floor is absolute and protocol-bound.}
The zero-shot Qwen3-VL-8B router and GRPO directly from the 2B base
(both reward arms) \emph{never} emit a \tcall{}: across all 1{,}000 test
records, tool invocations are 0, consistent across three independent
configurations; a loose substring re-scan finds 3 ``T\_call''
occurrences, all in the E12 arms' evaluation outputs (intrinsic rows),
where emitting a call is the taught behavior, while all
3{,}000 evaluation outputs give 0 under a permissive regex,
excluding parser false negatives (E3a, E3b). Any post-training extrinsic
competence must therefore be trained in, not elicited: zero-emission
under the \tcall{} protocol is a \emph{format-compliance} behavior
(\cref{sec:whysft}), not a capability ceiling.

\paragraph{Alignment SFT unlocks extrinsic routing (largest
single gain).} Alignment SFT lifts extrinsic routing from $0\%$ to
\textbf{80.75\%} (95\% Wilson CI $[0.766,0.843]$) (323/400), with 398 emissions and tool selection 0.905.

\paragraph{$R_{\mathrm{WA2}}$ GRPO suppresses and stabilizes.} On top of
alignment SFT: extrinsic routing holds at 80.75\%; intrinsic routing
$0.970\to\mathbf{0.985}$ (mis-emissions on intrinsic tasks reduced);
emissions $398\to384$; tool selection $0.905\to\mathbf{0.914}$
($+0.9$ points on overall routing, i.e., the 9 suppressed intrinsic
mis-emissions; extrinsic tool selection unchanged at 323/400); overall
accuracy $0.2430\to\mathbf{0.2500}$
($+0.7$ points). The suppression reaches answer quality: intrinsic
answer accuracy rises (classification $0.435\to0.460$, reasoning
$0.435\to0.450$).

\paragraph{Pairwise McNemar tests on the headline gains.}
All checkpoint comparisons are paired on the same 1{,}000 records, and
we report exact pairwise McNemar tests. E13$\to$E15: extrinsic tool
selection is identical at 323/400 ($p=1.0$ by construction) and the Mean
column moves $0.2430\to0.2500$ (exact McNemar $p=0.35$): \emph{no
significant gain}. The GRPO stage's contribution is behavioral
stabilization and the suppression of 9 intrinsic mis-emissions (calls
$398\to384$; intrinsic routing $0.970\to0.985$), not an accuracy gain
(the remaining deltas lie within the noise band). E15$\to$E17 (the
two-turn model under its strict-return protocol,
\cref{sec:conversion}): overall $p=9.5\times10^{-52}$ and extrinsic
$p=2.8\times10^{-54}$ (both on answer accuracy under the strict-return
protocol) are highly significant gains, while intrinsic
$p=7.0\times10^{-11}$ quantifies the significant degradation of intrinsic
routing ($0.985\to0.913$), the measured price of the repair. The
emission unlock $0\to323/400$ is significant against
any zero-call baseline at $p$ of order $2^{-323}$ (trivially so). The
single-seed limitation is \cref{sec:limitations}. Point estimates on
the full $n{=}1000$ carry ${\pm}2.7$pt 95\% Wilson sampling noise.

\paragraph{Trivial routing baselines.} On the same per-record routing
aggregate (mean over 1{,}000 records), the trivial policies score:
always-call 0.400, never-call 0.600, and the E19 text-only arm 0.348,
against E15's 0.914. The trained router beats the best trivial policy
(never-call) by $0.314$ (trivial policies score identically under both
calibers since they never choose a wrong tool; the E19 text-only arm's
0.348 is tool-selection scored), and the text-only prompt variant is worse than
always-calling, confirming the visual-grounding result
(\cref{sec:pmimage}).

\paragraph{The native-channel zero-shot baseline brackets the problem.}
A full-registry zero-shot run through the official \texttt{tools=}
channel (Hermes parser; base 2B, all 1{,}000 records, greedy) shows the
opposite failure. The model emits on 395/400 extrinsic records (98.75\%)
but also on 591/600 intrinsic records (suppression 1.5\%), with
near-zero tool-name accuracy (intrinsic 0.015, extrinsic 0.12, mean
routing 0.057; 986 calls): emission without trained routing is
worse than never calling. Together the two zero-shot rows bound the
problem: the \tcall{}-protocol baseline shows zero emission, the
native-channel baseline shows emission without discrimination; training
supplies the missing discrimination (0.914).

\paragraph{A note on the Mean column.} The zero-shot overall
accuracy (0.2630) is \emph{higher} than E15's (0.2500): intrinsic
templates are easy for the zero-shot model (600/600, see
\cref{tab:routing}), while the E13/E15 extrinsic arm, though routing
correctly 80.75\% of the time, has final answers suppressed by the
observation$\to$answer conversion gap (detection 0.045 / segmentation
0.005; \cref{sec:extrinsic}). The value of two-stage training is
therefore measured by the \emph{routing} metrics (Extr.\ / T-sel.\ /
Calls); Mean is a full-set reference only.

\begin{figure}[t]
  \centering
  \includegraphics[width=0.9\textwidth]{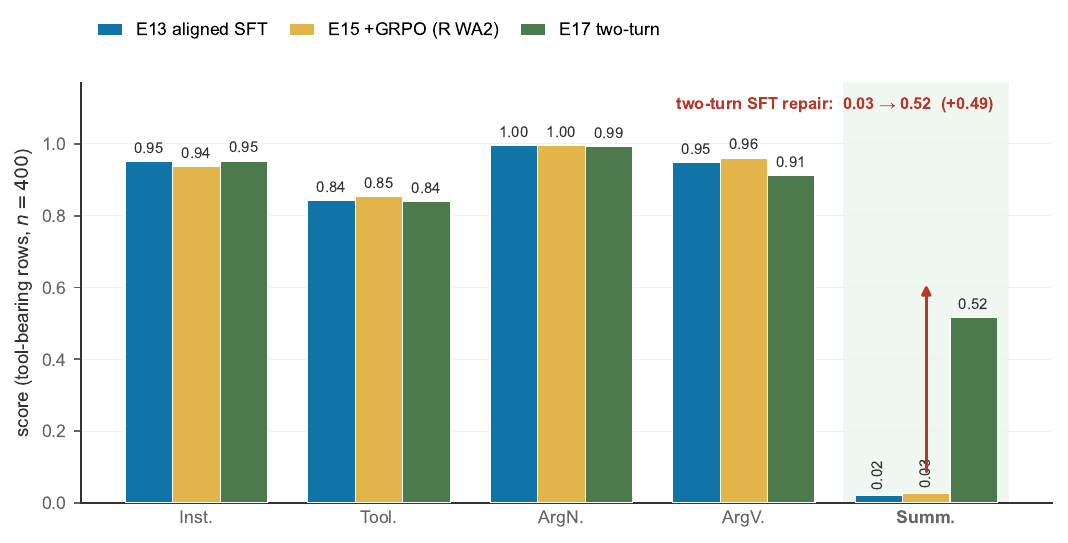}
  \caption{Five-fold decomposition of the extrinsic pipeline on the 400
  tool-bearing records (\cref{tab:pertask}). Argument folds are
  near-perfect while final answers stay broken for E13/E15: the
  layer-4 break is in converting tool text into answers; the two-turn
  SFT (E17) repairs exactly that fold.}
  \label{fig:decomp}
\end{figure}

\paragraph{Where the answers break: parameters are right, conversion is
broken.} \cref{tab:pertask} (\cref{fig:decomp}) decomposes the extrinsic
pipeline into five
folds: a parseable action is emitted (Inst), the tool is correct
(Tool), required argument names are present (ArgN), argument values are
correct (ArgV), the final answer is correct (Summ). For E13/E15 the
picture is unambiguous: argument folds are near-perfect (ALL ArgV
0.951/0.963) while final answers are broken (detection 0.035/0.045,
segmentation 0.005); \emph{the break is in converting tool text into
answers, not in the calling itself}. The two-turn trajectory SFT of
\cref{sec:twoturn} flips the picture under the real (strict-return)
protocol: tool-bearing Summ $0.020\to0.518$ (\cref{sec:conversion}).

\begin{table}[t]
\centering\footnotesize\setlength{\tabcolsep}{4pt}
\caption{Five-fold decomposition (Inst / Tool / ArgN / ArgV / Summ) of
the extrinsic pipeline, computed on the needs-tool records: detection
and segmentation rows on the 200 records of each family, ALL on the
400 tool-bearing records. Fold denominators: Inst/ArgN/ArgV/Summ over
the 400 needs-tool records; Tool over the emitted-call records
(denominators 380\,/\,375\,/\,381 for E13/E15/E17). (Overall test
accuracy, a different quantity, appears in \cref{tab:routing}.) E17 is
the two-turn model of \cref{sec:twoturn} under the real (strict-return)
protocol of \cref{sec:conversion}. Answers produced without any tool
call (\texttt{answer\_wo\_tool\_call}): E13 $=$ 20, E15 $=$ 25, E17 $=$
19; E17's 381 emitted-call records are exactly those with a single
well-formed call (\texttt{single\_call\_ok}). Family rows are computed
within the decomposition pipeline over $n{=}200$ per family; the ALL row
is the routing-pipeline tool-selection score over 380 evaluated rows
(20 execution-failure exclusions), so family and ALL rows use different
denominators by construction.}
\label{tab:pertask}
\begin{tabular}{llccccc}
\toprule
Model & Family & Inst & Tool & ArgN & ArgV & Summ \\
\midrule
Aligned SFT (E13) & detection & 0.905 & 0.685 & 0.994 & 0.909 & 0.035 \\
 & segmentation & 0.995 & 1.000 & 0.997 & 0.990 & 0.005 \\
 & ALL & 0.950 & 0.850 & 0.996 & 0.951 & 0.020 \\
\midrule
\quad $+$ GRPO $R_{\mathrm{WA2}}$ (E15) & detection & 0.880 & 0.705 &
0.994 & 0.932 & 0.045 \\
 & segmentation & 0.995 & 1.000 & 0.997 & 0.990 & 0.005 \\
 & ALL & 0.938 & 0.861 & 0.996 & 0.963 & 0.025 \\
\midrule
Two-turn SFT (E17, real) & ALL & 0.953 & 0.840 & 0.992 & 0.913 &
\textbf{0.518} \\
\bottomrule
\end{tabular}
\end{table}

\begin{figure}[t]
  \centering
  \includegraphics[width=0.92\textwidth]{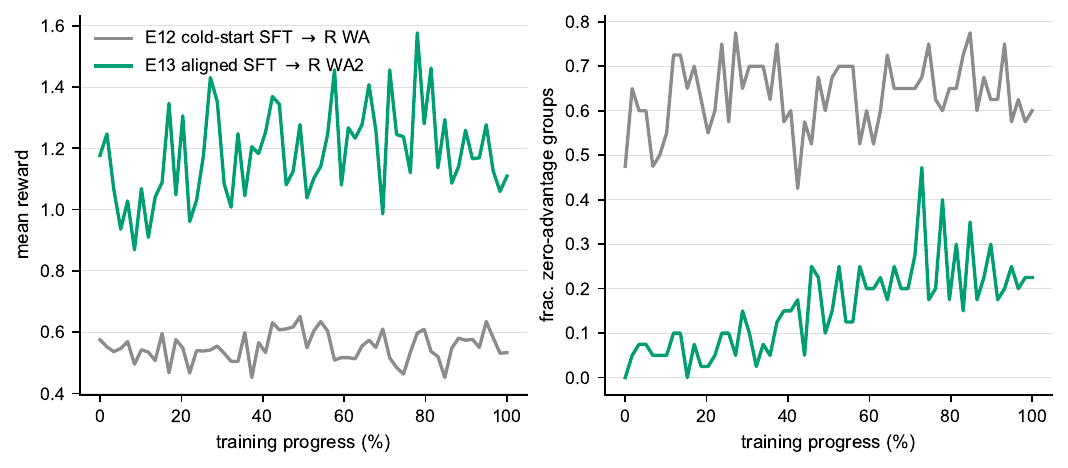}
  \caption{GRPO training dynamics (per-log-interval values, $x$ = training
  progress). On the gray curve (E12: cold-start SFT then
  $R_{\mathrm{WA}}$ GRPO), the
  reward stays low and most rollout groups have zero advantage, giving
  no gradient signal for emission. After alignment SFT (E15, green), the
  reward starts high and zero-variance groups stay an order of magnitude
  lower: emission is seeded, and the reward can shape it.}
  \label{fig:grpo}
\end{figure}

\paragraph{Training dynamics: each recipe step is visible on the curve.}
\cref{tab:dynamics} and \cref{fig:grpo} compare three training
configurations. GRPO from the
base model produces zero gradient on 85--90\% of groups; an unaligned SFT
cold start lowers zero-variance to ${\approx}0.43$--$0.78$ (median
${\approx}0.6$), but completion length
converges to direct-answer length (the $R_{\mathrm{WA}}$ reward
economics, \cref{sec:whyreward}); aligned SFT $+$ $R_{\mathrm{WA2}}$
starts at reward 1.18, keeps zero-variance groups at $0$--$0.1$ early
and at ${\le}0.3$ for most of training (peak 0.47, median
${\approx}0.15$), an order of magnitude below both unaligned arms,
with sequence lengths stable at \tcall{}-trajectory length: emission
behavior is seeded, aligned, and locked in by the reward structure.

\begin{table}[t]
\centering\footnotesize\setlength{\tabcolsep}{3pt}
\caption{Training dynamics of three configurations. Zero-variance groups
are rollout groups whose four rewards are identical (zero advantage, zero
gradient); mean length is the token length of generated completions.
Ranges are min--max over logged steps of the respective runs.}
\label{tab:dynamics}
\begin{tabular}{lp{2.9cm}p{4.0cm}p{3.8cm}}
\toprule
Metric & GRPO from base (E7a) & SFT cold start $+$ GRPO
$R_{\mathrm{WA}}$ (E12) & Aligned SFT $+$ GRPO $R_{\mathrm{WA2}}$ (E15) \\
\midrule
Initial reward & ${\approx}$0.058 & ${\approx}$0.55 (first-6: 0.5455) &
\textbf{1.18} \\
\texttt{frac\_reward\_zero\_std} & 0.85--0.9 & 0.43--0.78 &
\textbf{0--0.1 early, ${\le}0.3$ for most of training (max 0.47)} \\
Completions mean length (tokens) & 59--69 & 11--27 (converges to direct
answers; the $R_{\mathrm{RA}}$ arm collapses to 8.5--9.7) &
\textbf{165--180 (stable)} \\
\bottomrule
\end{tabular}
\end{table}

\subsection{Compression-Interface Characterization and Component
Ablations}
\label{sec:ablation}

\paragraph{Paired characterization on three benchmarks.} \cref{tab:main}
measures the compression interface against its 2048-px thumbnail baseline
under matched greedy decoding on identical stratified samples
(Qwen3-VL-8B). The paired accuracy deltas are $-1.0$ (LRS-VQA), $-3.4$
(MME-RS) and $-2.9$ (XLRS-lite) points, all negative, while the
board's token cost is fixed and does not grow with source size
(evaluation images: median side 6{,}044\,px, maximum 27{,}328\,px). This
is precisely the positioning basis of \cref{sec:evidence}: between
``fixed cost, no local detail (thumbnail)'' and ``local detail at a cost
growing with size (high-resolution crop stitching)'', the interface
offers a type-adaptive middle ground at a fixed budget, with a
$1.31\times$ detail ceiling (\cref{sec:ceiling}). \weaveearth{}'s
paper-reported full-set numbers (33.38/47.38/47.14) use a different
protocol and are never placed in the same row. Per-benchmark note:
MME-RS shows the largest drop ($-3.4$), with the position subset at
$-10.9$ points paired, consistent with the lost-global-continuity
mechanism (\cref{sec:ceiling}).

\begin{table}[t]
\centering\small\setlength{\tabcolsep}{3pt}
\caption{Compression-interface paired characterization (Qwen3-VL-8B,
greedy, identical 1{,}500-record stratified samples per benchmark;
accuracy \%). The evidence board's token cost is fixed (thumbnail $+$
board, two views ${\approx}5$k tokens) and does not grow with source
size.}
\label{tab:main}
\begin{tabular}{lccc}
\toprule
Benchmark & Thumbnail 2048 (interface off) & Thumbnail $+$ board
($k{=}6$) & Paired $\Delta$ \\
\midrule
LRS-VQA (1{,}500 strat.) & \textbf{33.20} & 32.20 & $-1.0$ \\
MME-RealWorld-RS (1{,}500 strat.) & \textbf{39.33} & 35.93 & $-3.4$ \\
XLRS-Bench-lite (1{,}500 strat.) & \textbf{41.47} & 38.60 & $-2.9$ \\
\bottomrule
\end{tabular}
\end{table}

\paragraph{Component ablations.} \cref{tab:ablation} reports the five
variants on the same $n{=}500$ stratified subset, greedy, $k{=}6$,
following the \weaveearth{} protocol. As measured: on static
VQA the full system (0.2940) does not exceed any leave-one-component
variant (0.3000--0.3140); the components make \emph{no positive
contribution} on this task distribution. The total range is $\le 2.0$
points, within the $\pm1$--2-point greedy noise band, and we make no
significance claims about single-point orderings; the shape nonetheless
agrees with \cref{tab:main}, jointly supporting the compression-interface
positioning. Methodological note: a full-system row must come from the
same subset as the ablated rows (mixing sample sizes, e.g.\ $n{=}1500$
vs.\ $n{=}500$, folds sampling noise into every delta); the E4\_full
same-subset control row exists for this reason.

\begin{table}[t]
\centering\small
\caption{Component ablations on LRS-VQA following the \weaveearth{}
protocol (accuracy; $n{=}500$ stratified, greedy, $k{=}6$; the
full-system row is the E4\_full same-subset control).}
\label{tab:ablation}
\begin{tabular}{lccccc}
\toprule
Variant & \gcc{} & \mses{} & \sem{} & \tpeb{} & Accuracy \\
\midrule
w/o \gcc{} & -- & \checkmark & \checkmark & \checkmark & 0.3040 \\
w/o \mses{} & \checkmark & -- & \checkmark & \checkmark & \textbf{0.3140} \\
w/o \sem{} & \checkmark & \checkmark & -- & \checkmark & 0.3080 \\
w/o \tpeb{} & \checkmark & \checkmark & \checkmark & -- & 0.3000 \\
\method{} (full) & \checkmark & \checkmark & \checkmark & \checkmark & 0.2940 \\
\bottomrule
\end{tabular}
\end{table}

\paragraph{Evidence-budget scan.} \cref{tab:kscan} scans
$k \in \{2,4,6,8,10\}$ on the same subset: the curve is flat (no
rise-then-fall shape; maximum range 1.6 points, within the noise band),
and the $k$ pipeline itself is verified correct by the budget probe
(\cref{sec:evidence}). For the compression interface, the budget does
not change static-VQA performance within $2 \le k \le 10$; the choice of
$k$ should be driven by the target task type's detail demand and the
token budget, not by an accuracy curve.

\begin{table}[t]
\centering\small
\caption{Evidence-budget scan ($n{=}500$ same-subset, greedy).
$^{*}k{=}6$ is the full-system same-subset control (E4\_full).}
\label{tab:kscan}
\begin{tabular}{lccccc}
\toprule
$k$ & 2 & 4 & 6 & 8 & 10 \\
\midrule
Accuracy & 0.3080 & 0.3040 & 0.2940$^{*}$ & \textbf{0.3100} & 0.3020 \\
\bottomrule
\end{tabular}
\end{table}

\subsection{Efficiency}
\label{sec:efficiency}

\begin{table}[t]
\centering\small
\caption{End-to-end latency. Hardware and protocol differ across rows and
are stated per row; paper-anchored rows are quoted from the respective
papers and serve as order-of-magnitude references only. Relative times
are normalized to our single-pass pipeline.}
\label{tab:efficiency}
\begin{tabular}{lccc}
\toprule
Method & Hardware / protocol & s / sample & Relative \\
\midrule
Thumbnail baseline 2048 (official script) & 4090, $n{=}300$, greedy &
2.84 & 0.39$\times$ \\
Zoom baseline (ours, E6, multi-round) & 4090, $n{=}50$ & 4.33 &
0.59$\times$ \\
Single-pass evidence construction (ours, E1a, greedy) & 4090, $n{=}1500$ &
\textbf{7.31} & 1$\times$ \\
WeaveAgent E15 single-pass (VagueUHR-test) & 4090, $n{=}1000$ & 6.17 &
0.84$\times$ \\
\weaveearth{} (paper-reported) & A100 & 7.59 & 1.04$\times$ \\
ZoomSearch (paper-reported, multi-round search) & --- & 54.25 &
7.4$\times$ \\
\remoteagent{} (paper-reported, non-UHR) & 8$\times$4090 & 1.18 & --- \\
\bottomrule
\end{tabular}
\end{table}

Single-pass evidence construction costs 7.31\,s per sample on a
consumer RTX\,4090 (2.57\,s retrieval $+$ 4.74\,s inference, greedy;
\cref{tab:efficiency}), on
par with \weaveearth{}'s reported 7.59\,s on an A100 and about
$7.4\times$ faster than reported multi-round visual search (ZoomSearch,
54.25\,s per sample; vs.\ reported numbers, cross-protocol). The trained
agent's own single-pass cost on the \vagueuhr{}-test split is 6.17\,s
per record (E15: 2.32\,s retrieval $+$ 3.85\,s inference, $n{=}1{,}000$,
4090); the two-turn extrinsic path adds one more generation round on top
of this figure, which we did not time separately. The
two-stage architecture's efficiency follows
from its design (\cref{sec:overview}): routing is a single text-first
decision stage (the $\pm$-image ablation quantifies the visual share of
the unified-prompt routing cost at $0.257$\,s text-only vs.\ $3.85$\,s;
\cref{sec:pmimage}), evidence construction is single-pass, and the tool
path executes in a single round; there is no multi-round zoom-and-answer
loop. For reference, the zoom baseline we implemented (E6, 4.33\,s on
its 50-record protocol) is cheaper per sample than the single-pass
pipeline but operates in the zoom-and-answer regime whose reported
analog costs an order of magnitude more, and it does not bound the tool
path, which never pays the visual-compression cost at all. Latency rows
are deliberately \emph{not} paired with an accuracy comparison against
zoom-and-search methods: they neither perform tool routing nor report
results under the \vagueuhr{} protocol, so no like-for-like accuracy
row exists; the table is an order-of-magnitude cost reference
only.

\subsection{Query-Register Gradient: Clear, Template, and
LLM-Rewritten}
\label{sec:register}\label{sec:wording}

\begin{figure}[t]
  \centering
  \includegraphics[width=0.9\textwidth]{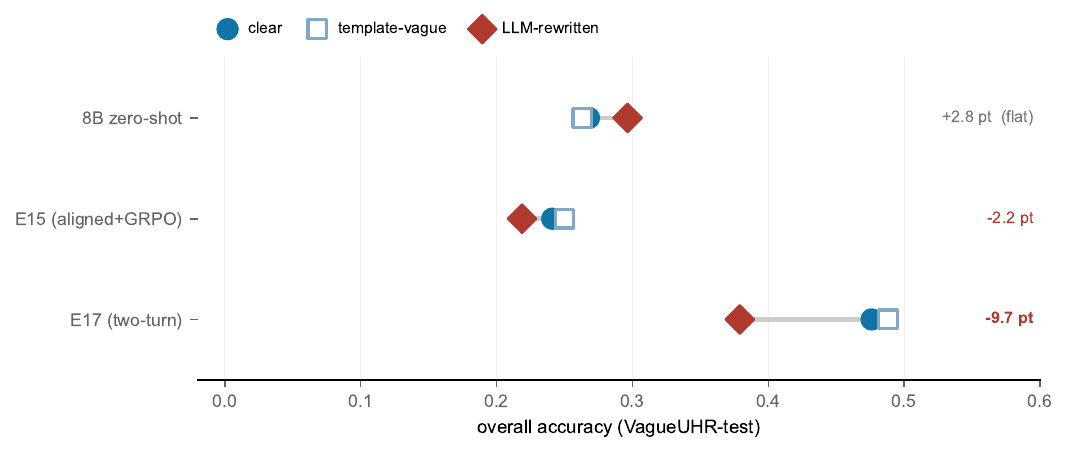}
  \caption{The query-register gradient: overall accuracy for three
  checkpoints under three query registers (filled circles: clear; open
  squares: template-vague; diamonds: LLM-rewritten; per-cell extrinsic
  routing in \cref{tab:register} below). The untrained row is flat: a
  zero-shot model cannot route, so register does not matter. Training
  creates a register gradient: template-vague queries behave like clear
  ones ($\le$1.2 points), while LLM-rewritten queries cost the trained
  checkpoints $2$--$11$ points. The vagueness axis is real, measured,
  and partially absorbed by the deeper recipe (E17 exceeds E15 in every
  register).}
  \label{fig:register}
\end{figure}

Because Stage-A's decision interface is text-first by design (a
premise the $\pm$-image ablation of \cref{sec:pmimage} tests directly),
robustness to query surface wording is critical for the
architecture. \cref{fig:register} visualizes the register matrix;
\cref{tab:register} arranges the wording evidence as a
$3\times3$ register matrix and separates three gradients;
\cref{tab:wording} reports the untrained row's underlying paired
measurement.
\textbf{(1) The untrained row is register-blind.} The zero-shot 8B
model's extrinsic emission stays at the floor in all three registers
($0$\,/\,$0$\,/\,$0.25\%$): the zero-shot register insensitivity is a
routing-capability gap, not a wording effect, and the native-channel
baseline shows it is not even an emission gap (\cref{sec:routing});
the register axis only becomes visible once alignment training makes
emission possible at all. The paired measurement (E14) applies the E3a
configuration to the 992 record-by-record paired template queries and
the LLM-visual-rewrite split, zero-shot Qwen3-VL-8B on both arms:
natural rewrites are, if anything, slightly \emph{easier} (overall
accuracy $0.2651\to0.2964$, $+3.12$ points, McNemar exact $p=0.0433$
(95 vs.\ 126 discordant pairs);
per-type movement a reshuffle, Figure~S1),
and the absolute
floor survives rewriting (template arm: 0 tool calls; rewritten arm: 2
calls, 1 extrinsic routing hit among 399 extrinsic records). The
zero-emission baseline of \cref{sec:routing} is a property of the
model--protocol pair, not of template phrasing.

\begin{table}[t]
\centering\small
\caption{Query-register gradient on \vagueuhr{}-test: overall accuracy
(primary) and extrinsic routing (secondary) for the untrained 8B row and
the two aligned checkpoints under three query registers. The E17 row
uses the two-turn protocol with Eval-A oracle-grounded tool returns
(return quality is an upper bound; the strict cross-mode protocol itself
is not the performance upper bound, which is Eval~B,
\cref{sec:conversion}); all LLM-rewritten cells use
the 992-record paired rewritten split, clear/template cells the full
1{,}000-record test set. Det./Seg.\ give per-type accuracy on the
rewritten register (8B from the E14 arm of \cref{tab:wording}); the
two-turn repair's largest per-type gains are exactly in these columns.}
\label{tab:register}
\begin{tabular}{lcccccccc}
\toprule
& \multicolumn{2}{c}{Clear} & \multicolumn{2}{c}{Template-vague} &
\multicolumn{2}{c}{LLM-rewritten} & \multicolumn{2}{c}{LLM per-type} \\
\cmidrule(lr){2-3}\cmidrule(lr){4-5}\cmidrule(lr){6-7}\cmidrule(lr){8-9}
Checkpoint & overall & extr & overall & extr & overall & extr & det &
seg \\
\midrule
8B zero-shot & 0.268 & 0.000 & 0.263 & 0.000 & \textbf{0.296} & 0.0025 &
0.231 & 0.405 \\
E15 (aligned) & 0.241 & 0.800 & 0.250 & 0.8075 & 0.219 & 0.667 & 0.085 &
0.005 \\
E17 (two-turn) & 0.476 & 0.7875 & 0.488 & 0.800 & 0.379 & 0.659 &
0.382 & 0.235 \\
\bottomrule
\end{tabular}
\end{table}

\textbf{(2) Template vagueness is not a difficulty axis for
the trained rows.} Template-mode vagueness only adds tone words and
never deletes task-defining words, and the trained checkpoints move
accordingly little: E15 $0.241\!\to\!0.250$, E17
$0.476\!\to\!0.488$ (clear$\to$template; $|\Delta|\le1.2$ points).
\textbf{(3) The rewritten register is where wording bites, and the cost
is now quantitative.} On the LLM-rewritten column E15 loses $3.1$ points
overall ($0.250\!\to\!0.219$) and $14$ points of extrinsic routing
($0.8075\!\to\!0.667$); E17 loses $10.9$ points overall
($0.488\!\to\!0.379$) and again $14$ points of extrinsic routing
($0.800\!\to\!0.659$). The two-turn repair largely survives the
colloquial register; E17 stays $16$ points above E15 on the same
column ($0.219\!\to\!0.379$), with the largest per-type gains exactly
where the repair targets them (detection $0.085\!\to\!0.382$,
segmentation $0.005\!\to\!0.235$), but the routing give-back returns
in sync: conversion robustness and wording robustness are in measurable
tension.

\begin{table}[t]
\centering\small
\caption{Query-wording sensitivity, paired across splits (992 records;
Qwen3-VL-8B zero-shot, greedy). Per-type $n$: 197\,/\,199\,/\,199\,/\,197\,/\,200
(classification/counting/detection/reasoning/segmentation). The template
arm reproduces the E3a
configuration on the paired subset; the LLM arm uses the visually
rewritten queries. $\Delta = +3.12$ points, McNemar exact $p = 0.0433$.
$^{*}$With (near-)zero emissions the T-sel.\ aggregate reduces to
counting correct no-calls on intrinsic records, a bookkeeping
artifact, not a measured capability (\cref{tab:routing}).}
\label{tab:wording}
\begin{tabular}{lccccc}
\toprule
Split & Overall acc. & Intr.\ routing & Extr.\ routing & T-sel. & Calls \\
\midrule
Template (992 paired) & 0.2651 & 1.000 & 0.000 & 0.598$^{*}$ & 0 \\
LLM-rewritten (992) & \textbf{0.2964} & 0.998 (592/593) & 0.0025
(1/399) & 0.598$^{*}$ & 2 \\
\bottomrule
\end{tabular}
\end{table}

\subsection{The Stage-A $\pm$-Image Ablation: Emission Suppression Is
Visually Grounded}
\label{sec:pmimage}

\begin{figure}[t]
  \centering
  \includegraphics[width=0.98\textwidth]{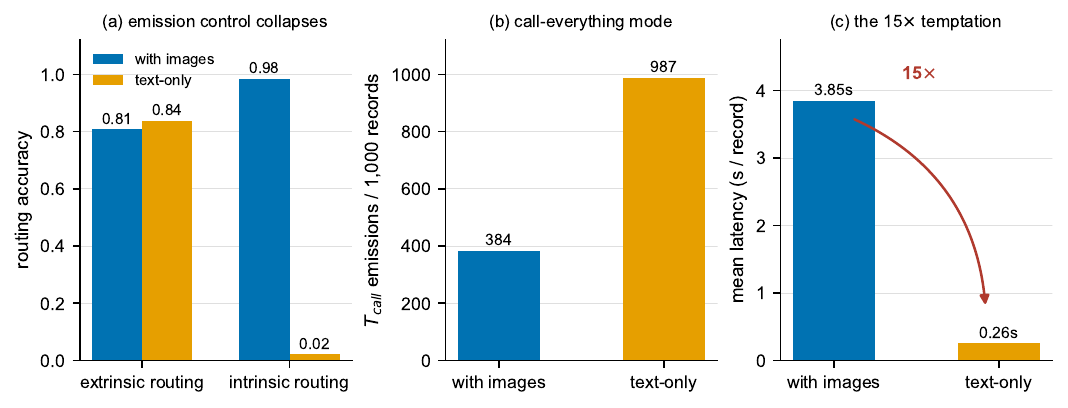}
  \caption{The $\pm$-image ablation (E19, E15 checkpoint, 1{,}000
  records, identical prompts). Removing image tokens keeps extrinsic
  routing (0.838 vs.\ 0.808) but collapses intrinsic routing to 0.022:
  the model emits a call on almost every record (987 vs.\ 384).
  Emission suppression, knowing when \emph{not} to call, is
  visually grounded. The $15\times$ latency gap makes a truly text-only
  router an attractive but unattained target.}
  \label{fig:pmimage}
\end{figure}

E19 measures the Stage-A design basis directly (\cref{fig:pmimage}): the
E15 checkpoint
evaluates all 1{,}000 test records twice, with prompts and records
identical and only the image tokens removed from the input (text-only
arm vs.\ image-conditioned arm, the latter being the E15 main evaluation
of \cref{sec:routing}); the description-block text is retained in the
no-image arm. The contrast is one-sided. Extrinsic routing is
\emph{preserved} without images ($0.8375$ vs.\ $0.8075$, $+3.0$ points),
but intrinsic routing \emph{collapses} ($0.985\!\to\!0.0217$): the
text-only policy emits a \tcall{} on 987 of 1{,}000 records (vs.\ 384),
and tool selection degrades to $0.348$ (vs.\ $0.914$). Per-record
routing latency drops $15\times$ ($0.257$\,s vs.\ $3.85$\,s): the
visual context dominates the cost of the unified prompt, so a genuinely
text-only router would be dramatically cheaper. The reading: the strong
form of the text-only design basis (\cref{sec:router}) does not hold for
the trained policy. The visual input is the signal that
\emph{suppresses} emission on intrinsic queries, so knowing when
\emph{not} to call is visually grounded in UHR imagery, while the
emission format and extrinsic discrimination survive on text alone. A
truly text-only router therefore remains an unattained target; distilling
one from the image-conditioned policy is future work
(\cref{sec:limitations}).

\subsection{Tool-Execution Conversion: Oracle Attribution and Two-Turn
Repair}
\label{sec:conversion}

\cref{sec:routing} localized the extrinsic bottleneck to the
observation$\to$answer conversion. This subsection (i) attributes it
via an oracle experiment (E16) and (ii) repairs it with the two-turn
SFT of \cref{sec:twoturn} (E17). \emph{Protocol disclosure}:
all tool returns in this subsection are oracle/annotation-grounded, not
live expert services; no number here is an end-to-end measurement
against real deployed backends, and the two E17 protocol points below
must be read as an upper bound and a cross-mode generalization point
respectively. A deployment attempt of the real SkySense/RemoteSAM
service stack on the single-GPU evaluation box was recorded
(mmdet/mmcv and service weights absent;
\texttt{T1\_live\_backend\_report.json}); live end-to-end evaluation
remains future work, and the oracle-protocol qualification of every
number in this subsection stands.

\begin{figure}[t]
  \centering
  \includegraphics[width=0.85\textwidth]{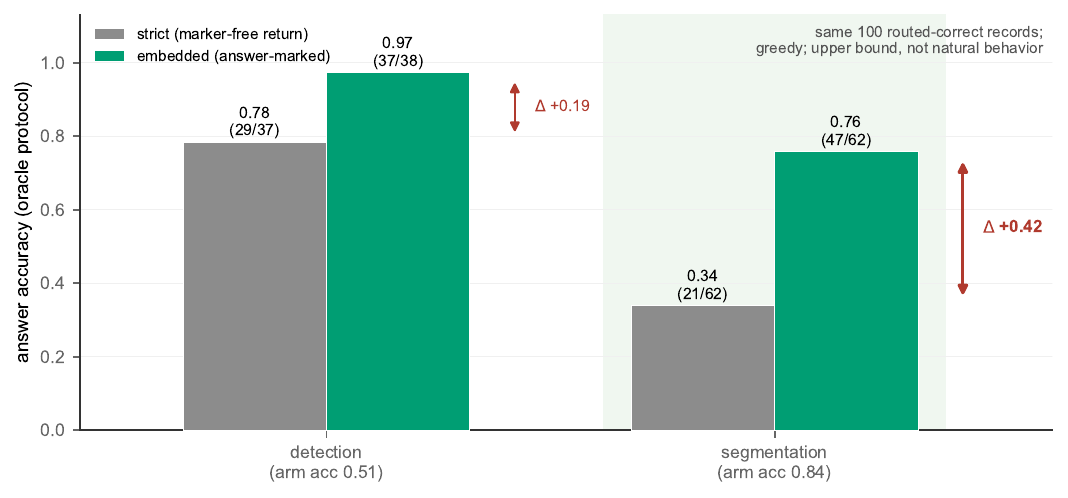}
  \caption{Oracle attribution on the 100 routed-correct records
  (\cref{tab:oracle}): answer accuracy when the tool return is
  marker-free (strict) vs.\ answer-marked (embedded). The gap
  concentrates in segmentation, whose mask-type returns carry almost no
  textual information. Upper bound, not natural behavior.}
  \label{fig:oracle}
\end{figure}

\begin{figure}[t]
  \centering
  \includegraphics[width=0.92\textwidth]{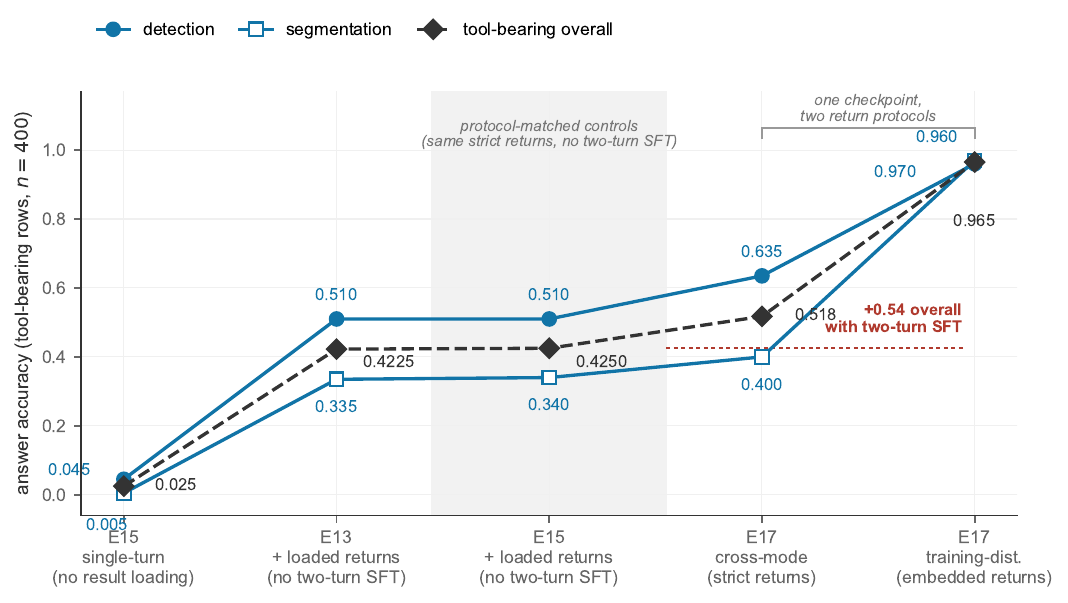}
  \caption{The layer-4 repair, end to end. Answer accuracy on the 400
  tool-bearing records across the five-point ladder: E15
  (single-turn, no result loading), the two protocol-matched control
  arms (E13/E15 with the same strict returns loaded but no two-turn
  training), E17 under marker-free cross-mode returns, and E17 under
  training-distribution (embedded) returns. The ladder branches at the
  last two points: one trained checkpoint, two return protocols (the
  intrinsic task types and routing metrics are identical across them).}
  \label{fig:threepoint}
\end{figure}

\paragraph{E16: oracle attribution.} On the E15 checkpoint (aligned SFT
$+$ $R_{\mathrm{WA2}}$ GRPO), 100 routed-correct records stratified by
task type (37--38 detection / 62 segmentation; two-round protocol,
greedy), the
\emph{strict} arm strips answer markers from tool returns (N/A counts
filled with ground-truth numbers so returns look like real detector
output), while the \emph{embedded} arm stuffs the \remoteagent{}
trajectory text with answers in place. \cref{tab:oracle}
(\cref{fig:oracle}): converting
marker-free tool text into answers is hard for a 2B model: overall
$0.505$ ($n{=}100$, 95\% Wilson CI $[0.41,0.60]$) strict vs.\ $0.840$ embedded, and the gap concentrates in
segmentation ($0.339$ vs.\ $0.758$), whose mask-type returns are
inherently uninformative as text; at $n{=}100$ the arm-level gap
(${\approx}0.34$) dwarfs the ${\pm}0.1$ binomial noise of either
estimate. Routing itself is not the issue
(no-call: 1/100 strict, 0/100 embedded). By the pre-registered gate
(choose \emph{embedded} iff strict accuracy falls more than $0.10$
below embedded), embedded is selected as the E17 training-stuffing
format. Both arms are oracle upper bounds, not natural system
performance. These 100 records are routed-correct records, a
self-selected and easier population than the full 400 tool-bearing rows,
so E16 is an attribution probe, not a protocol-matched baseline for
Eval~A.

\begin{table}[t]
\centering\small
\caption{E16 oracle attribution (100 routed-correct records stratified
by task type: 37--38 detection / 62 segmentation; two-round protocol,
greedy; oracle
returns). ``Strict'' strips answer markers from tool returns;
``embedded'' stuffs \remoteagent{}-style trajectory text with answers in
place. The detection denominator under strict is 37 (one no-call
record).}
\label{tab:oracle}
\begin{tabular}{lcccc}
\toprule
Arm & Detection & Segmentation & Overall & No-call \\
\midrule
Strict (marker-free) & 0.784 (29/37) & 0.339 (21/62) & 0.505 & 1/100 \\
Embedded (answer-bearing) & \textbf{0.974} (37/38) & \textbf{0.758}
(47/62) & \textbf{0.840} & 0/100 \\
\bottomrule
\end{tabular}
\end{table}

\paragraph{E17: two-round evaluation under two protocols.} The two-turn
model is evaluated on the full 1{,}000-record test set under two
protocols that must be read together (\cref{tab:twoturn}). \emph{Eval~A
(strict-fallback oracle protocol)}: the tool service was unreachable, so every record fell
back to strict marker-free returns (the fallback rate is 1.0);
because training used embedded returns, this is the
\emph{cross-mode generalization} point. \emph{Eval~B (embedded
protocol)}: returns stuffed in the training-distribution embedded
format, the \emph{training-distribution upper bound}. Under Eval~A,
extrinsic answer accuracy reaches $0.635$ (detection) and $0.400$
(segmentation) despite the train/test return-format mismatch; the
repair generalizes across return modes rather than memorizing a format.
Under Eval~B, $0.960$/$0.970$. The intrinsic task types and all routing
metrics are identical across the two evals, which is exactly the
protocol
sanity check one wants, since intrinsic records involve no tool
returns. The five-point ladder on extrinsic answer accuracy (needs-tool
Summ; \cref{fig:threepoint}): $0.025$ (E15, single-turn, no result
stuffing) $\to$ $0.4225$/$0.4250$ (E13/E15 $+$ loaded strict returns, no
two-turn training) $\to$ $0.518$ (E17, cross-mode strict returns) $\to$
$0.965$ (E17, training-distribution informative returns).
Protocol-matched control arms separate the two ingredients: with the
same strict returns loaded but no two-turn training, E13 reaches
$0.4225$ and E15 $0.4250$ on the 400 tool-bearing rows: the
observation loading accounts for the bulk of the jump
($0.025\to0.425$), and the two-turn SFT stage adds a further $+9.25$
points to $0.5175$ (McNemar $b=21$, $c=58$, $p=3.8\times10^{-5}$;
against the E13 base $p=3.2\times10^{-5}$). The two controls are
statistically indistinguishable from each other ($p=1$): GRPO
contributes nothing to conversion, by design. Detection improves
$0.510\to0.635$ and segmentation $0.340\to0.400$ under the same pairing,
with no per-type regression against the matched controls. The costs: extrinsic routing gives back a little
($0.8075\to0.800$), tool selection $0.914\to0.868$, and calls rise to
433 (vs.\ 384 for E15; 398 for E13); intrinsic routing reads $0.913$.
(Relative to the E15 single-turn specialist: since E17 branches from the
E13 base, this delta reflects the two-turn branch's training data and
objective rather than the marginal effect of two-turn SFT on top of
GRPO.)

\begin{table}[t]
\centering\footnotesize\setlength{\tabcolsep}{3pt}
\caption{E17 two-round evaluation under two protocols (full
1{,}000-record test set, greedy; oracle returns in both, no live
service). Eval~A: oracle-grounded strict marker-free returns via
per-record fallback (fallback rate 1.0, disclosed); cross-mode
generalization point. Eval~B: embedded returns, training distribution;
upper bound. Intrinsic types and routing metrics are identical across
evals (protocol sanity). Class./Count./Reas./Det./Seg.\ = answer
accuracy per task family (200 records each); Extr./Intr.\ = routing
accuracies; T-sel.\ = tool selection. Control arms: same strict-return
protocol and oracle results, no two-turn training; E15$\to$E17
difference on the same 400 rows is $+9.25$pt (McNemar
$p=3.8\times10^{-5}$).}
\label{tab:twoturn}
\begin{tabular}{lcccccccccc}
\toprule
Protocol & Overall & Class. & Count. & Reas. & Det. & Seg. & Extr. &
Intr. & T-sel. & Calls \\
\midrule
Eval A (strict fallback) & 0.488 & 0.560 & 0.330 & 0.515 & 0.635 &
0.400 & 0.800 & 0.913 & 0.868 & 433 \\
Eval B (embedded) & \textbf{0.667} & 0.560 & 0.330 & 0.515 &
\textbf{0.960} & \textbf{0.970} & 0.800 & 0.913 & 0.868 & 433 \\
\midrule
E13 $+$ loaded strict returns (control) & --- & --- & --- & --- & 0.510
& 0.335 & \multicolumn{4}{c}{extr.\ ans.\ $0.4225$} \\
E15 $+$ loaded strict returns (control) & --- & --- & --- & --- & 0.510
& 0.340 & \multicolumn{4}{c}{extr.\ ans.\ $0.4250$} \\
\bottomrule
\end{tabular}
\end{table}

\begin{figure}[!htbp]
  \centering
  \includegraphics[width=0.98\textwidth]{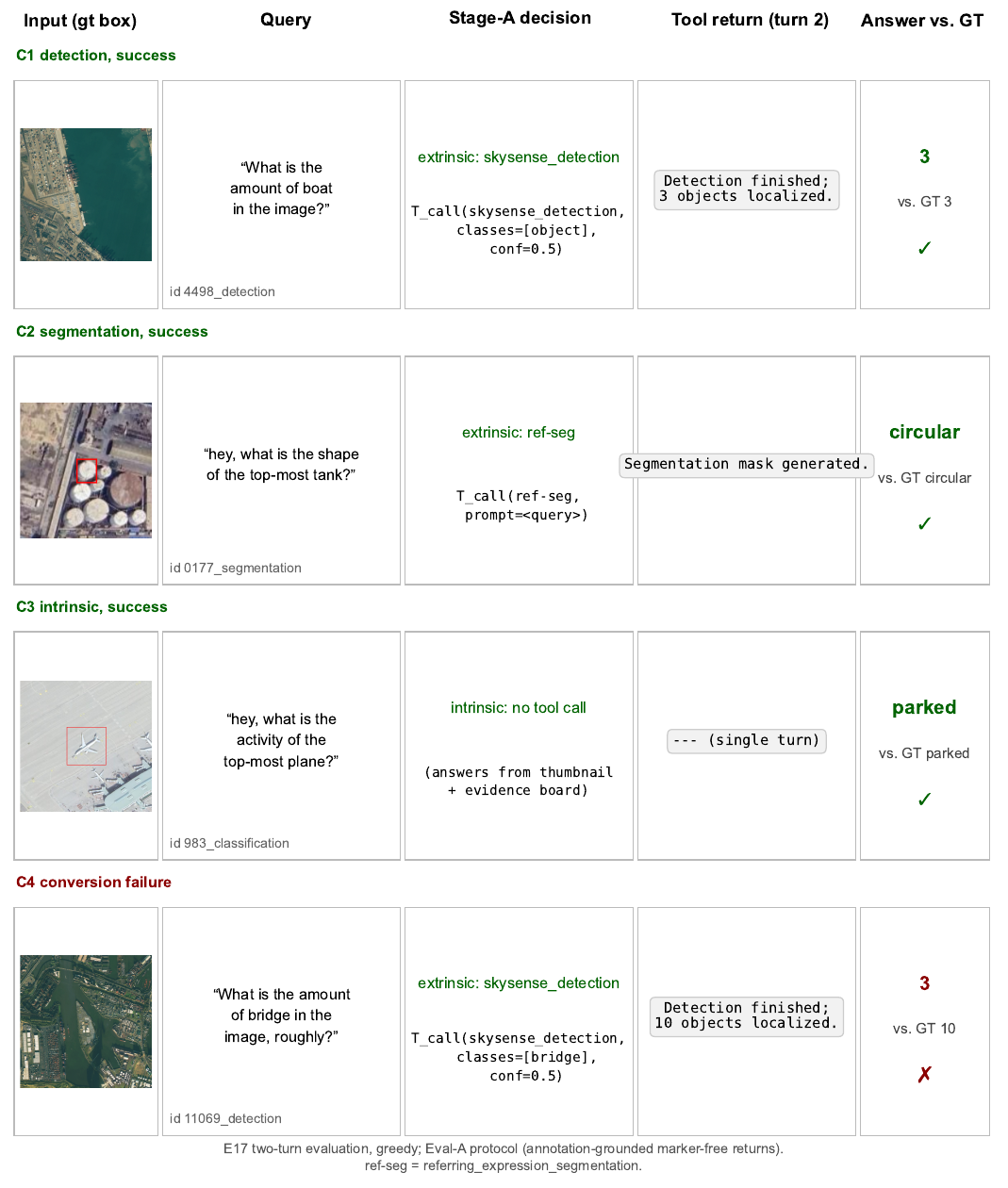}
  \caption{Four qualitative cases from the E17 two-turn evaluation
  (Table~S7; greedy, Eval-A protocol).
  Each band shows the input
  crop (ground-truth box in red where defined), the query, the Stage-A
  decision with the emitted \tcall{}, the marker-free turn-2 tool
  return, and the final answer against the ground truth. C4 is the
  instructive conversion failure: the count is present in the return
  (``10 objects localized'') yet the model answers 3.}
  \label{fig:casesfig}
\end{figure}

\subsection{Qualitative Analysis}
\label{sec:qualitative}

\cref{fig:casesfig} walks through the two-stage pipeline on four
selected records; the full case table with question IDs
(Table~S7) is provided in Appendix~C for exact
cross-checking. C1 and C2 show the extrinsic path end to end: Stage-A's
router emits a well-formed \tcall{}, the observation is loaded as a
second user message, and the answer follows. C3 shows the intrinsic
path on a vagueness-marked query: the router correctly refrains from
calling a tool and the model answers directly. C4 is the instructive
failure: the return already contains the count (``10 objects
localized''), yet the model answers 3. This is a numeric-\emph{extraction}
failure, not a routing or tool failure; it is the residual mode
behind the strict-protocol detection ceiling ($0.635$,
\cref{sec:conversion}), while under the training-distribution returns
of Eval~B the same conversion is learned (detection $0.960$ there).
C2's tool return, moreover, carries no shape information at all
(``Segmentation mask generated.''), so the correct answer can only come
from the visual channel; this is exactly the return-informativeness gap
that \cref{sec:conversion} measures with the E16 strict/embedded arms
(segmentation $0.339$ vs.\ $0.758$), and it motivates reading mask-type
returns through a visualized overlay rather than raw text in future
work. Finally, the evaluation protocol skips records with retrieval
errors rather than scoring them as wrong; the skip-on-retrieval-error
clause never fired in any reported run (zero skips), and the E17 run
contained no such record, so no retrieval-bottleneck case survives
into the figure; that phenomenon is documented separately in the
compression-interface analysis (\cref{sec:ceiling}).

\section{Design Rationale and Analysis}
\label{sec:rationale}

The recipe of \cref{sec:training} raises two ``why'' questions that must
be answered with data: why alignment SFT is necessary
(\cref{sec:whysft}) and why the incentive must be routing-first
(\cref{sec:whyreward}). A third subsection characterizes the boundary of
the compression interface (\cref{sec:ceiling}); a fourth records
implementation differences and prompt-format sensitivity as neutral
protocol-alignment notes (\cref{sec:repro}). Each item is one paragraph
with data. All statements about \weaveearth{}~\cite{ma2026weaveearth}
are neutral records of protocol differences, not criticisms.

\subsection{Why Alignment SFT Is Necessary: Two Mechanisms}
\label{sec:whysft}

\paragraph{Mechanism 1: the exploration deadlock (GRPO cannot create
behavior from nothing).} GRPO is group-relative: advantage
$A = (r - \bar r)/(\mathrm{std} + \epsilon)$. When the policy never
emits a tool call, all four rollouts in a group are direct answers,
rewards are identical, $\mathrm{std}=0$, all advantages are 0, and the
gradient is zero. Measured GRPO from the 2B base: initial reward
${\approx}0.058$, \texttt{frac\_reward\_zero\_std} $= 0.85$--$0.9$:
85--90\% of groups produce no learning signal. The zero-shot
zero-emission floor (top rows of \cref{tab:routing}) is this mechanism's
direct projection: GRPO can only amplify variance in behavior that
already exists. Competing hypotheses were excluded one by one: a loose
regex re-scan of all raw evaluation outputs finds 0/4{,}000 containing
\texttt{T\_call} (not a parser false negative); the median tool-call
completion is $p50 = 193$ characters ($p90 = 229$, ${\approx}70$--80 tokens), below
max\_new\_tokens $= 128$ (not truncation); the evaluation prompt is
confirmed to contain the 15-line tool catalog (not a missing catalog).
Corroborating evidence from the native channel: with the official
\texttt{tools=} channel (Hermes parser), the \emph{base}
Qwen3-VL-2B-Instruct emits 20/20 calls on a 20-task smoke set but hits
the correct tool name in only 4/20: zero emission under the custom
\tcall{} protocol is a \emph{format mismatch}, and tool
\emph{selection} is the skill that actually needs training. The
full-set native-channel baseline confirms this at scale: 98.75\%
extrinsic emission (395/400) with mean routing 0.057
(\cref{sec:routing}). Seeding by
SFT is therefore required, matching ToRL~\cite{li2025torl} (forced tool
calls seed exploration), ToolRL~\cite{qian2025toolrl}, and
DAPO~\cite{yu2025dapo} (zero-variance pathologies). Measured: an SFT
cold start lifts emission from 0 to 0.55 (500 steps) $\to$ 0.65
(1{,}000 steps) and the GRPO initial reward to ${\approx}0.55$, with
zero-variance groups dropping to 0.43--0.78 (\cref{tab:dynamics}).

\paragraph{Mechanism 2: format binding (emission is bound to the
prompt distribution).} The cleanest causal evidence comes from the
unaligned SFT: with the \emph{same} E12 merged weights and the same tool
tasks, the training-format prompt elicits a 0.65 emission rate while the
evaluation-format prompt elicits 0/400 (the evaluation path does inject
the real evidence board, ruling out ``evidence not given''). Byte-level
comparison finds $\ge$10 non-image differences between the training and
evaluation prompts (full listing: Appendix~A,
Table~S1);
the fatal one is the compound effect of the last-line format
instruction: an evaluation prompt that moves the decision sentence to
mid-prompt and appends a final line supporting only direct answers is,
to a 2B model, a different task. The zero-shot model's 0\% emission is
the prior version of the same mechanism.

\paragraph{Design conclusion.} Alignment SFT, which rebuilds the
cold-start data with evaluation-identical prompts and image formats
(E13), is the only path from $0\%$ to 80.75\% extrinsic routing.
Methodological moral: any ``behavior disappeared after SFT'' phenomenon
should trigger a training/evaluation prompt-distribution consistency
audit before model capability is blamed. This is what ``alignment''
means in the recipe's first step.

\subsection{Why the Incentive Must Be Routing-First: Reward Economics}
\label{sec:whyreward}

\paragraph{$R_{\mathrm{WA}}$'s structure makes direct answering
expected-optimal.} From Table~S4 ($w_t = 0.3$, no-call
consolation $0.2$): a correct call on an extrinsic task caps at $0.35$,
a correct direct answer scores $1.11$, and calling on intrinsic tasks
(73\% of training data, 2{,}400/3{,}273) scores $0$; aggregating
(\cref{tab:rewardexpect}), always answering directly ($0.11 +
p_{\mathrm{ans}}$) beats correct routing ($0.35$) once
$p_{\mathrm{ans}} \ge 0.24$:
the RL-rational policy is ``treat tool tasks as answer tasks''.
Measurements agree: under E12-$R_{\mathrm{WA}}$ mean completion length
converges to 16--25 tokens (i.e.\ no calls), and under
$R_{\mathrm{RA}}$ RL \emph{actively removes} emission (a bare \tcall{}
has no \texttt{<answer>} span and scores a constant zero; mean length
collapses to 8.5--9.7 tokens); even SFT-seeded behavior is washed
out by a wrongly structured reward.

\paragraph{$R_{\mathrm{WA2}}$ inverts the expectation.} The three
structural modifications (delete the no-call consolation score
($0.2\to0$), raise the routing weight ($0.3\to1.0$), and split the
format term into two $0.5$ components) invert the expectations: correct
direct answers actually rise ($1.11\to1.25$), but correct routed calls
rise more ($0.35\to2.25$), and correct routing $2.25$ $\gg$ direct answering
$0.25$--$1.25$ per case (a correct route is credited with the answer
term automatically); the inversion is parameter-free, holding
for any answer accuracy (\cref{tab:rewardexpect}). E15 measurements
(\cref{tab:dynamics}): initial reward \textbf{1.18}, zero-variance
groups an order of magnitude below the unaligned arms, mean length
stable at \textbf{165--180}, emission locked in by the reward
structure; on the evaluation side, tool selection
$0.905\to\mathbf{0.914}$ ($+0.9$ points on overall routing, i.e., the 9
suppressed intrinsic mis-emissions; extrinsic tool selection unchanged
at 323/400) and overall accuracy
$0.2430\to\mathbf{0.2500}$ (\cref{sec:routing}).

\paragraph{Design conclusion.} $R_{\mathrm{WA2}}$'s value shape is
``maintain and suppress'' rather than ``raise further'': under the
single-turn evaluation protocol, most of the tool-selection gain is
already claimed by alignment SFT (80.75\%); $R_{\mathrm{WA2}}$
stabilizes emission on the evaluation distribution, eliminates
intrinsic mis-emissions, and squeezes out a further $+0.9$ points of
tool selection. Without alignment SFT the reward-design difference is
invisible (the two E12 arms tie at evaluation); alignment and
incentive must be repaired together, which is why the recipe's two
steps are inseparable.

\subsection{The Boundary of the Compression Interface: the
$1.31\times$ Ceiling and Conditions of Applicability}
\label{sec:ceiling}

\paragraph{The ceiling.} The \tpeb{} gives each evidence cell a fixed
448-px patch on a $6\times6$ grid; relative to the 2048-px full-scene
thumbnail, the effective detail gain of a selected region is
$448 \times 6 / 2048 \approx 1.31\times$, \emph{constant regardless
of the source image's resolution}: a 27{,}328-px scene earns the same
31\% as a 4{,}096-px one. The board also covers only 6 of 36 cells (a
sixth of the field of view) and destroys global spatial continuity. Two
mechanisms explain the full negative shape of
\cref{tab:main,tab:ablation}: (i) tasks that depend on whole-scene
layout do not benefit: MME-RS position questions lose $-10.9$ points
paired, because local detail cannot compensate for lost global geometry;
(ii) retrieval is recall-bounded: if the \gcc{} scorer
(Eq.~\eqref{eq:gcc}) does not rank the relevant region into the top-8
anchors, the downstream VLM never sees it, and the budget-$k$ selection
(Eq.~\eqref{eq:mses}) only compounds this by design. A retrieval audit
over the 660
spatial-ground-truth test records (gt-box-center criterion) measures
recall@8 anchors $= 0.291$ and recall@6 final board $= 0.230$:
71\% of queried targets are never reachable through the evidence
board, and MSES compression discards a further 6 points; per task
type the pattern is uniform (0.20--0.32 across types, per-type detail
in Appendix~C; counting, $n{=}2$, has no spatial ground truth and
is excluded). Query--patch
similarity retrieval is weakest exactly where UHR tasks are hardest
(whole-scene counting, spatial relations, change detection).

\paragraph{Interventional repairs (harder boundary evidence than
observation).} After the three-benchmark paired negatives, we
systematically tried four families of repairs, five configurations in
total (paired, greedy, seed-0 stratified 300
records, baseline $=$ interface off): routing spatial questions to the
global view (MME $+1.35$, $p=0.65$; LRS $-2.1$, $p=0.41$, with reasoning
$-9.3$ from ``where is'' misrouting); removing the thumbnail for detail
questions with $k{=}4$ large cells (MME $-4.7$, $p=0.08$, color
$-10.1$); keeping the thumbnail with $k{=}4$ native-resolution cells
(MME $+0.34$ / LRS $-1.7$; shape $+2.8$ but count $-10.3$); and a
four-quadrant coverage guarantee (neutral). \emph{None significantly
beats the thumbnail baseline (all $p \in [0.08, 1.0]$), and directions
contradict each other.} The mechanism explanation is consistent: the
board's detail advantage is locked at ${\approx}1.31\times$ by the
448-px budget; raising per-cell resolution necessarily reduces cell
count, and the coverage loss cancels the gain; ``spatial questions to
the global view'' is the only stably positive component
(${\approx}2$ points).

\paragraph{Design conclusion.} No positive component contribution on
static VQA (\cref{tab:ablation}), a flat $k$ curve (\cref{tab:kscan}),
and no significant repair together support the positioning of
\cref{sec:evidence}: \emph{\tpeb{}$+$\sem{} is a fixed-budget
compression interface whose benefit lies in token determinism and
local-structure preservation for fine-grained task types, not in
average accuracy}. Enable it per task type (fine-grained grounding),
not by default.

\subsection{Implementation Differences and Prompt Sensitivity
(Neutral Records)}
\label{sec:repro}

This subsection records differences between the \weaveearth{} released
implementation and the paper's algorithm, and prompt-format
sensitivity, as neutral protocol-alignment records; Table~S5
(Appendix~B) collects the measurements, and
Figure~S2 (Appendix~C) plots the
Category-line control.
(i) Under the paired greedy protocol the released script's own gain is
$+2.67$ points (29.67\% $\to$ 32.33\%; 89 vs.\ 97 of 300 correct, 24 vs.\ 32 discordant pairs,
McNemar exact $p = 0.350$) versus the paper-reported $+6.70$
(26.68\% $\to$ 33.38\%) on the full set; the paper's baseline
configuration (prompt, decoding) is not documented, so we record the
protocol gap and treat the paper and the release as two distinct
objects. (ii) The released \mses{} selection truncates the candidate
pool to the top-8 anchors before the budget greedy, so the greedy loop
never executes; we reimplement per the paper's algorithm
(\cref{sec:evidence}) and audited all shared components item by item.
(iii) A Category task-type line present only in the treatment arm had
been inferred, by cross-protocol inference, to contribute
${\approx}3.7$ points; the paired control on the same 300 samples
(60 per task type, E15 checkpoint, greedy) gives $0.2667$ vs.\ $0.2833$,
$\Delta = -1.7$ points \emph{with} the line, McNemar exact
$p = 0.51$, per-type movement a reshuffle; the Category line therefore
does not leak answer information and the ${\approx}3.7$-point inference
is refuted. (iv) The mere order of MCQ option instructions versus the
\sem{} metadata block swings accuracy by ${\approx}10.5$ points,
silently (quarantined self-audit, indicative; 34\% of outputs degrade
to prose that fails every matching rule); we therefore adopt two rules
throughout: byte-identical prompt templates across baseline and
treatment arms except for the mechanism under test, and McNemar tests
on paired binary outcomes rather than point deltas alone.

\section{Limitations}
\label{sec:limitations}

\begin{enumerate}[leftmargin=*,itemsep=1pt]
  \item \textbf{Backbone scale.} 8B GRPO does not fit a 24-GB GPU, so all
  two-stage training and evaluation use Qwen3-VL-2B (the zero-shot floor
  row of \cref{tab:routing} is 8B). Scaling the alignment pipeline to 8B
  (alignment SFT $\rightarrow$ $R_{\mathrm{WA2}}$ GRPO re-run at 8B,
  requiring ${\ge}48$ GB) and re-measuring the routing table at that
  scale remains future work.
  \item \textbf{A truly text-only router remains unattained.} The
  image-free decision stage would run at $15\times$ lower latency
  ($0.257$\,s vs.\ $3.85$\,s per record); the trained policy does not
  yet operate in this regime (\cref{sec:pmimage}), and distilling a
  genuinely text-only router is future work.
  \item \textbf{Wording robustness is now measured for the aligned
  checkpoints, but register coverage is partial.} The register-gradient
  matrix (\cref{tab:register}, \cref{sec:register}) extends the paired
  E14 measurement from the 8B zero-shot floor to both aligned checkpoints
  across the template and LLM-rewritten registers; multi-turn, real-user
  phrasing remains unmeasured.
  \item \textbf{The conversion repair is not end-to-end on real
  services.} E16/E17 use oracle/annotation-grounded returns; Eval~A's
  fallback rate is 1.0. Tool experiments mix real deployments
  (DirectSAM-class) with annotation-grounded simulated backends,
  disclosed per table; a deployment attempt of the
  SkySense\slash RemoteSAM service stack on the evaluation box is
  recorded (mmdet/mmcv and service weights absent;
\cref{sec:conversion}), and end-to-end success against live expert
backends is not yet measured; the routing give-back (tool selection
$0.914\to0.868$; calls $384\to433$) is the price of the repair.
  \item \textbf{$R_{\mathrm{WA2}}$'s routing-first shape carries a
  call-versus-answer tension.} Calling correctness and answer
  correctness diverge in the data (\cref{tab:pertask}: ArgV
  $0.951/0.963$ vs.\ Summ $0.020/0.025$ for E13/E15), and E17's
  conversion repair buys answer accuracy at a routing give-back (tool
  selection $0.914\to0.868$, intrinsic routing $0.985\to0.913$;
  \cref{sec:conversion}). This two-dimensional trade-off is disclosed
  rather than resolved.
  \item \textbf{Unseen-tool generalization is negative: registration
  makes a tool visible, not usable.} All three trained checkpoints route
  0/62 on the training-unseen sm3det records, with emissions collapsing
  onto the seen tools (48\,/\,3\,/\,11 for E15; \cref{sec:unseen}). The test
  pool is three tools; routing generalization under a larger catalog is
  untested.
  \item \textbf{The region parameter $p$ of $\tcall(e_k, p)$ is not
  evaluated.} In the reported runs the tools consume the full image or
  the query text as the prompt; region grounding, the syntax of $p$, and
  a region-correctness metric are unmeasured, although acting on the
  right image region is the core UHR difficulty (a target may cover
  $0.14\%$ of the scene; \cref{fig:teaser}).
  \item \textbf{The tool pool is generic CV, not GIS.} Detection,
  segmentation and counting backends only; no GIS operations (region
  statistics, overlay, buffering, area/distance measurement) and no use
  of CRS, GSD or georegistration. Absent bi-temporal sources, the
  change-detection shape is collapsed to detection (\cref{sec:splits}),
  so temporal reasoning is outside the evaluated scope; a geospatial
  workflow case is future work.
  \item \textbf{Training queries are templated.} Template-mode vagueness
  is narrower than open-ended user phrasing (\texttt{vague\_level} all
  $0.5$); the LLM rewrite channel exists but is not enabled
  for training.
  \item \textbf{Prompt fragility is a deployment risk.} The E12 audit
  shows that one prompt-side edit can silently destroy trained routing:
  the same weights emit on $0.65$ of tool tasks under the training-format
  prompt but $0/400$ under the evaluation-format prompt
  (\cref{sec:whysft}).
  \item \textbf{Benchmark contamination and geographic concentration.}
  Test records derive from public benchmarks (LRS-VQA, MME-RealWorld-RS,
  XLRS-Bench), so pretraining exposure of the base VLM cannot be
  excluded, and the sources' geographic coverage is limited (the flagged
  1213 cluster; \cref{sec:honesty}).
  \item \textbf{Statistical and deployment scope.} Except for the
  300-record official comparison, all benchmark numbers are stratified
  estimates; sampling noise is quantified only through paired tests, and
  all runs use a single seed (seed 0).
  XLRS: 7 of 8 subcategories (SR unavailable; OP 53.9\% of our sample).
\end{enumerate}

\section{Conclusion}
\label{sec:conclusion}

\method{} is a two-stage tool-routing agent for ultra-high-resolution
remote sensing. \emph{Stage~A} is a routing-first decision stage: the
$\pm$-image ablation (E19) shows its emission suppression is visually
grounded (removing image tokens keeps extrinsic routing but collapses
intrinsic routing to 0.022), and it performs
intrinsic/extrinsic discrimination and tool
selection from the query and the catalog, on the design basis established by the
text-routing literature. \emph{Stage~B} executes conditionally: the
intrinsic path answers from the thumbnail with an optional fixed-budget
evidence-compression interface; the extrinsic path executes \tcall{} on
the original full-resolution imagery and answers from tool observations
along two-turn trajectories with observation masking. On the
\vagueuhr{} dual-split corpus, the alignment-SFT $\rightarrow$
$R_{\mathrm{WA2}}$ GRPO recipe yields its checkable gain at the
alignment stage: alignment SFT lifts extrinsic routing from $0\%$
(zero-shot: zero emissions) to 80.75\% (323/400) with tool selection
0.905; the $R_{\mathrm{WA2}}$ GRPO stage leaves extrinsic tool selection
unchanged (323/400), suppresses 9 intrinsic mis-emissions (calls
$398\to384$, intrinsic routing $0.970\to0.985$; answer accuracy
classification $0.435\to0.460$, reasoning $0.435\to0.450$), and moves
overall accuracy $0.2430\to0.2500$ (within the noise band), with
healthy training dynamics (zero-variance groups an order of magnitude below
the unaligned arms). The evidence board is
positioned as a fixed-budget compression interface: two views
of ${\approx}5$k tokens at constant cost regardless of source size,
with a $1.31\times$ detail ceiling and three-benchmark paired deltas of
up to $-3.4$ points; single-pass evidence construction
runs at 7.31\,s per sample on a consumer 4090, on par with reported
A100 numbers and faster than reported multi-round
visual search (cross-hardware reference). Where the pipeline broke, we said so and
repaired what could be repaired: an oracle study attributed the
extrinsic answer gap to observation$\to$answer conversion (argument
values ${\approx}96\%$ correct, final answers 0.045), and extrinsic
answer accuracy rose from 0.025 to 0.425 by loading the observation
alone and 0.518 with the two-turn SFT stage ($p=3.8\times10^{-5}$),
under marker-free cross-mode returns (0.965 under training-distribution
returns), at a small routing cost. A
query-register gradient measurement (\cref{sec:register}) closes the
wording question for the aligned checkpoints: template-level vagueness
moves overall accuracy by at most $1.2$ points, while LLM-rewritten
phrasing costs $3.1$ (E15) and $10.9$ (E17) points overall with a
${\approx}14$-point extrinsic-routing give-back. Open problems
are equally explicit: 8B-scale
alignment, end-to-end conversion repair against live tool services,
register coverage beyond the two measured registers (multi-turn,
real-user phrasing), and a truly text-only router (E19 shows emission
suppression is visually grounded, so distilling one is future work).
Code, data,
and all evaluation protocols will be released.

\paragraph*{Funding.} No funding was received for this work.

\paragraph*{Declaration of competing interests.} The author declares
no competing interests.

\paragraph*{Data and code availability.} The \vagueuhr{} corpus
(query--trajectory splits for training and the dual-split test), all
per-record evaluation outputs, and the released code are openly
available (access details in the published version of this article;
imagery files are not redistributed --- they are obtained from the
source benchmarks). Construction scripts, prompt templates,
per-experiment configurations, and the remaining code components
accompany the full release.

\paragraph*{Declaration of generative AI use.} In accordance with the
Taylor \& Francis AI policy, the author discloses that large
language model assistants (GLM 5.3, Z.ai, via a coding agent) were used
during manuscript preparation for literature verification, development of
experiment and figure-generation code, language editing, and triage of
simulated peer-review reports. All scientific decisions, experiments,
numerical results, and conclusions are the author's own; every AI-assisted
output was reviewed and verified by the author, and no generative AI tool
was used to create scientific data or imagery.

\clearpage
\appendix

\setcounter{table}{0}
\renewcommand{\thetable}{S\arabic{table}}
\setcounter{figure}{0}
\renewcommand{\thefigure}{S\arabic{figure}}

\section{Training and Prompt Configuration}
\label{app:config}

This appendix collects the byte-level prompt-format audit underlying
Section~6.1 (\cref{tab:promptdiff}) and a register of every
experiment ID referenced in this paper, with its configuration and a
pointer to the table or figure that reports it
(\cref{tab:expregister}). The register lists facts only; no new claims
are introduced here.

\begin{table}[htbp]
\centering\footnotesize\setlength{\tabcolsep}{4pt}
\caption{Supplementary table (appendix A). Non-image differences between
the unaligned training prompt and the evaluation prompt (byte-level
audit; the evidence board is genuinely injected at evaluation time,
ruling out ``evidence not given'').}
\label{tab:promptdiff}
\begin{tabular}{cp{2.4cm}p{5.4cm}p{6.0cm}}
\toprule
\# & Difference & Training prompt & Evaluation prompt \\
\midrule
1 & Opening sentence & ``You are WeaveAgent answering\ldots'' & ``Answer
the question about\ldots'' \\
2 & Dual-image description block & none & thumbnail $+$ evidence-board
description \\
3 & Category line & none & \texttt{Category: <task\_type>} \\
4 & Query prefix & \texttt{User query: } & \texttt{Question: } \\
5 & Catalog position & before the query & after the query \\
6 & Decision-instruction position & last line, mentions both options &
mid-prompt, followed by 3 more blocks \\
7 & Decision-instruction wording & ``Either answer directly\ldots, or
emit a tool call\ldots'' (answer first) & ``If an external tool is
required\ldots instead of an answer'' (call as a conditional branch) \\
8 & Final format line & none (the decision sentence is last) & ``Return
only a short answer phrase: \texttt{<answer>...</answer>}'' (no
\tcall{} at all) \\
9 & Region metadata / \sem{} block & none (sem\_text presence 0/3{,}273)
& present \\
10 & Block separator & blank line & single newline \\
\bottomrule
\end{tabular}
\end{table}

\paragraph{\sem{} presence across prompt families.} \cref{tab:sempresence}
consolidates, from the disclosures of \cref{tab:promptdiff},
Section~3.5 and Section~6.1, and \cref{tab:cases}, where the
thumbnail, the \tpeb{} board, and the \sem{} block are present in each
prompt family used in this paper; no new measurements are introduced.

\begin{table}[htbp]
\centering\footnotesize
\caption{Supplementary table (appendix A). Presence of the thumbnail,
the \tpeb{} board, and the \sem{} block across prompt families
(consolidated from \cref{tab:promptdiff}, Sections~3.5 and~6.1, and \cref{tab:cases}; facts only, no new
measurements). E17 turn-1 is byte-identical to the E13 aligned
demonstrations, whose prompt and image formats are strictly identical
to evaluation; turn-2 loads the stuffed tool observation in the
\remoteagent{} inference format.}
\label{tab:sempresence}
\begin{tabular}{lp{2.7cm}p{2.7cm}p{3.4cm}}
\toprule
Prompt family & Thumbnail & \tpeb{} board & \sem{} block \\
\midrule
Routing-train prompts (E13/E15) & path (lazily decoded) & none & none
(sem\_text 0/3{,}273) \\
Main evaluation prompts & present & present (real injection) & present \\
E19 text-only arm (E15 ckpt.) & image removed; description text
retained & image removed; description text retained & retained as text
(byte-identical prompt) \\
E17 turn-1 / turn-2 & present (E13 demo images) & present
(evaluation-identical image format) & none (training side); turn-2 adds
the observation text \\
\bottomrule
\end{tabular}
\end{table}

\begin{table}[htbp]
\centering\footnotesize\setlength{\tabcolsep}{4pt}
\caption{Supplementary table (appendix A). Experiment register: one row
per experiment ID used in this paper --- configuration and result
pointer (facts only). ``Official 300'' is the released-script
300-record stratified paired replication.}
\label{tab:expregister}
\begin{tabular}{lp{7.2cm}l}
\toprule
ID & Configuration & Result \\
\midrule
E1a & Single-pass evidence pipeline, interface on (thumbnail $+$
board, $k{=}6$); three benchmarks, stratified 1{,}500 each, 8B, greedy;
timing anchor & Tables~6 and~9 \\
E1b & Evidence interface off (2048 thumbnail), same samples &
Table~6 \\
E3a & Zero-shot Qwen3-VL-8B routing evaluation & Table~3 \\
E3b & GRPO from the 2B base, $R_{\mathrm{RA}}$ / $R_{\mathrm{WA}}$
arms (merged E7a/E7b weights) & Table~3 \\
E4 / E4\_full & Component ablations with the same-subset full-system
control ($n{=}500$, greedy) & Table~7 \\
E5 & Evidence-budget $k \in \{2,\dots,10\}$ scan & Table~8 \\
E6 & Multi-round zoom baseline ($n{=}50$) & Table~9 \\
E7a / E7b & GRPO from base ($R_{\mathrm{RA}}$ / $R_{\mathrm{WA}}$);
training dynamics & Table~5 \\
E12 & Unaligned SFT cold start $+$ GRPO $R_{\mathrm{WA}}$; source of
the prompt audit & Table~5, \cref{tab:promptdiff} \\
E13 & Alignment SFT & Tables~3 and~4 \\
E14 & LLM-visual-rewrite wording arm (zero-shot 8B, 992 paired) &
Table~11, \cref{fig:wording} \\
E15 & Alignment SFT $+$ GRPO $R_{\mathrm{WA2}}$ & Tables~3, 4, and~5 \\
E16 & Oracle attribution, strict/embedded arms ($n{=}100$) &
Table~12 \\
E17 & Two-turn SFT; Eval-A / Eval-B protocols & Tables~13, 4, and~10 \\
E18 & Clear-register runs (E15 / E17 $\times$ clear) &
Table~10 \\
E19 & $\pm$-image ablation (E15 checkpoint, 1{,}000 records) &
Figure~7 \\
E20 & Category-line paired control (300 records, 60 per task type) &
\cref{tab:impldiff}, \cref{fig:e20} \\
Official 300 & Released-script stratified paired replication &
\cref{tab:impldiff} \\
\bottomrule
\end{tabular}
\end{table}

\section{Reward Scores and Neutral Implementation Records}
\label{app:impl}

This appendix collects the full $R_{\mathrm{WA}}$ per-case score table
(\cref{tab:rewardscores}; Sections~3.5 and~6.2) and the
neutral implementation-difference records of Section~6.4.

\begin{table}[htbp]
\centering\footnotesize
\caption{Supplementary table (appendix B). $R_{\mathrm{WA}}$ per-case
scores (measured configuration: $w_a{=}1.0$, $w_t{=}0.3$, $w_f{=}0.1$,
no-call consolation $0.2$). Total $= 1.0\cdot$answer $+\;
0.3\cdot$tool $+\; 0.1\cdot$format ($R_{\mathrm{WA}}$ weights); the
three columns are unweighted component scores: a binary indicator
(answer), a routing/consolation term (0.2 under $R_{\mathrm{WA}}$, 0
under $R_{\mathrm{WA2}}$), and the format term (0.5 per well-formed
component). On intrinsic tasks a no-call
earns the full tool term (there is no ground-truth tool); the structure
makes bare \tcall{} behavior --- exactly what SFT teaches --- nearly
worthless relative to direct answering.}
\label{tab:rewardscores}
\resizebox{\textwidth}{!}{%
\begin{tabular}{lcccc}
\toprule
Case & Answer indicator & Tool indicator & Format indicator & Total \\
\midrule
Intrinsic, direct answer \textbf{correct} & 1.0 & 1.0 & 0.5 & \textbf{1.35} \\
Intrinsic, direct answer wrong & 0 & 1.0 & 0.5 & 0.35 \\
Intrinsic, emits \tcall & 0 & 0.0 & 0.0 & \textbf{0.00} \\
Extrinsic, \textbf{correct} \tcall{} ($=$ SFT behavior) & 0 & 1.0 & 0.5 & \textbf{0.35} \\
Extrinsic, wrong \tcall & 0 & 0.0 & 0.5 & 0.05 \\
Extrinsic, no call $+$ wrong answer & 0 & 0.2 & 0.5 & 0.11 \\
Extrinsic, no call $+$ \textbf{correct} answer & 1.0 & 0.2 & 0.5 & \textbf{1.11} \\
Extrinsic, correct call $+$ correct answer (both required) & 1.0 & 1.0 & 1.0 & \textbf{1.40} \\
\bottomrule
\end{tabular}}
\end{table}

\begin{table}[htbp]
\centering\small
\caption{Supplementary table (appendix B). Implementation differences
and prompt sensitivity: neutral protocol-alignment records. The
MCQ-order row comes from a quarantined self-audit run and is indicative
only.}
\label{tab:impldiff}
\begin{tabular}{p{5.9cm}p{5.3cm}p{3.0cm}}
\toprule
Measurement & Numbers & Source \\
\midrule
\weaveearth{} paper-reported (LRS-VQA full set) & 26.68 $\to$ 33.38
($+6.70$) & paper \\
Released script, 300-record stratified paired, both arms greedy &
29.67 $\to$ 32.33 ($+2.67$, 24/32 discordant, McNemar exact $p=0.350$) & this work,
Section~5.1 \\
Category task-type line asymmetry (released baseline lacks it, treatment
arm has it) & paired control (E20): $\Delta=-1.7$ points, McNemar
$p=0.51$ --- no answer leakage; the earlier ${\approx}3.7$-point
cross-protocol inference is refuted & this work (E20) \\
MCQ option-instruction order (self-audit run, indicative) & 25.47 $\to$
35.93 (${\approx}10.5$-point swing; 34\% of outputs degrade to prose) &
quarantined audit \\
\bottomrule
\end{tabular}
\end{table}

\section{Supplementary Measurements and Qualitative Cases}
\label{app:cases}

This appendix collects supplementary figures and tables: the E14 per-task-type wording figure
(Section~5.5), the E20 Category-line control figure
(Section~6.4), the per-task-type GCC retrieval-recall
detail (\cref{tab:gccrecall}; Section~6.3), and the
qualitative case table of Section~5.8.

\begin{figure}[htbp]
  \centering
  \includegraphics[width=0.9\textwidth]{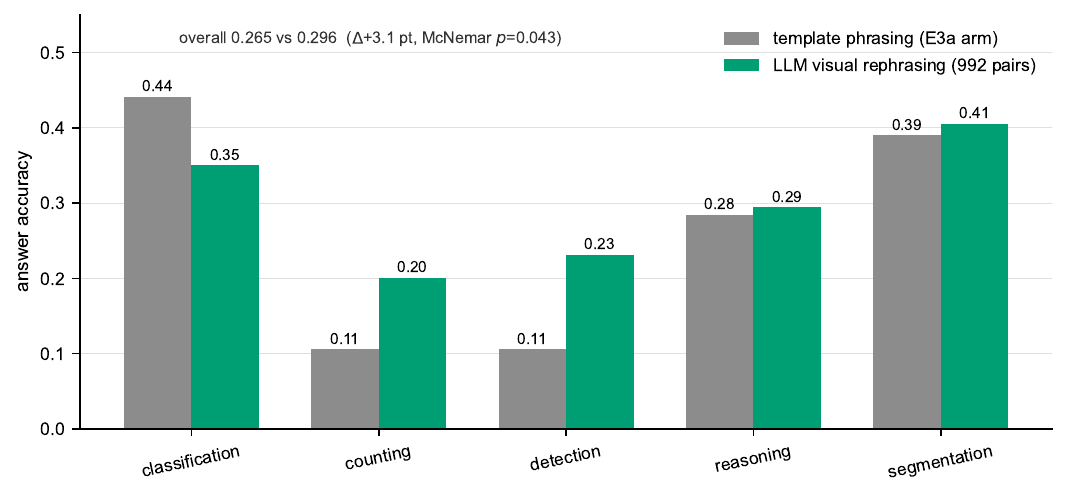}
  \caption{Supplementary figure (appendix C). Query-wording sensitivity
  on the 992 paired records (zero-shot Qwen3-VL-8B, greedy;
  Table~11): per-task-type accuracy on the template split
  vs.\ the LLM-visual-rewrite split. The movement is a per-type
  reshuffle rather than systematic damage.}
  \label{fig:wording}
\end{figure}

\begin{figure}[htbp]
  \centering
  \includegraphics[width=0.88\textwidth]{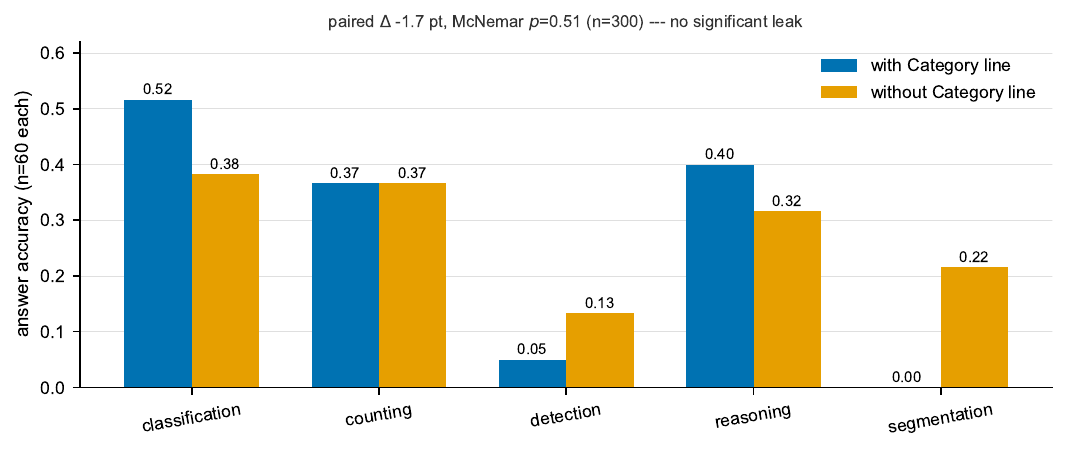}
  \caption{Supplementary figure (appendix C). Category-line paired
  control (300 records, 60 per task type, E15 checkpoint, greedy):
  per-type answer accuracy with vs.\ without the Category row
  (classification $+13.3$, detection $-8.3$, reasoning $+8.3$ points).
  The movement is a per-type reshuffle, not a systematic leak; the
  overall paired effect is $-1.7$ points ($p = 0.51$).}
  \label{fig:e20}
\end{figure}

\begin{table}[htbp]
\centering\small
\caption{Supplementary table (appendix C). GCC retrieval recall per
task type on the 660 spatial-ground-truth test records
(gt-box-center criterion; Section~6.3). Counting has no spatial
ground truth ($n{=}2$) and is excluded from the uniformity statement.
Classification and segmentation coincide exactly at this sample size
(57/200 at $k{=}8$ and 48/200 at $k{=}6$, identical for both types) --- verified against the
raw per-type output, not a transcription error.}
\label{tab:gccrecall}
\begin{tabular}{lccc}
\toprule
Task type & Recall@8 anchors & Recall@6 final board & $n$ \\
\midrule
Classification & 0.285 & 0.240 & 200 \\
Counting & 0.0 & 0.0 & 2 \\
Detection & 0.323 & 0.258 & 62 \\
Reasoning & 0.296 & 0.204 & 196 \\
Segmentation & 0.285 & 0.240 & 200 \\
\midrule
All spatial records & \textbf{0.291} & \textbf{0.230} & 660 \\
\bottomrule
\end{tabular}
\end{table}

\begin{table}[htbp]
\centering\footnotesize
\caption{Supplementary table (appendix C). Four qualitative cases from
the E17 two-turn evaluation (greedy; Section~5.7, Eval~A
protocol --- tool returns are annotation-grounded fallbacks,
marker-free). Question IDs are given for reproducibility.
\texttt{ref-seg} $=$ \texttt{referring\_expression\_segmentation}.}
\label{tab:cases}
\begin{tabular}{@{}p{1.1cm}p{3.4cm}p{4.4cm}p{3.6cm}p{2.6cm}@{}}
\toprule
Case & Query & Stage-A decision and \tcall{} & Turn-2 tool return & Output vs.\ GT \\
\midrule
C1 (det.) & ``What is the amount of boat in the image?'' (\texttt{4498})
& extrinsic: \texttt{T\_call(skysense\_detection, classes=[object],
confidence\_threshold=0.5)}
& ``Detection finished; 3 objects localized.''
& \textbf{3} vs.\ 3 \checkmark \\
\addlinespace[2pt]
C2 (seg.) & ``hey, what is the shape of the top-most tank?''
(\texttt{0177})
& extrinsic: \texttt{T\_call(ref-seg, prompt=$\langle$query$\rangle$)}
& ``Segmentation mask generated.''
& \textbf{circular} vs.\ circular \checkmark \\
\addlinespace[2pt]
C3 (intr.) & ``hey, what is the activity of the top-most plane?''
(\texttt{983})
& intrinsic: no call; answers from thumbnail (+ evidence board)
& --- (single turn)
& \textbf{parked} vs.\ parked \checkmark \\
\addlinespace[2pt]
C4 (fail) & ``What is the amount of bridge in the image, roughly?''
(\texttt{11069})
& extrinsic: \texttt{T\_call(skysense\_detection, classes=[bridge],
confidence\_threshold=0.5)}
& ``Detection finished; 10 objects localized.''
& 3 vs.\ \textbf{10} $\times$ \\
\bottomrule
\end{tabular}
\end{table}


\end{document}